%% file: main.tex
\PassOptionsToPackage{table}{xcolor}
\documentclass{article} 
\usepackage{iclr2027_conference,times}

\input{math_commands.tex}

\usepackage{hyperref}
\usepackage{url}

\usepackage[T1]{fontenc}
\usepackage[utf8]{inputenc}
\usepackage{microtype}
\usepackage{inconsolata}
\usepackage{graphicx}

\usepackage{fontawesome5}
\definecolor{darkblue}{rgb}{0, 0, 0.5}
\hypersetup{colorlinks=true, citecolor=darkblue, linkcolor=darkblue, urlcolor=darkblue}
\newcommand{\hficon}{\raisebox{-0.2em}
{\includegraphics[height=1.1em]{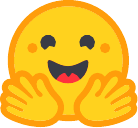}}}
\newcommand{\lnk}[1]{\makebox[1.6em][l]{#1}}

\usepackage{amsmath}
\usepackage{booktabs}
\usepackage{tabularx}
\usepackage{wrapfig}
\usepackage{subcaption}
\usepackage{algorithm}
\usepackage{algpseudocode}
\usepackage{xcolor}
\usepackage[most]{tcolorbox}  
\definecolor{rsnavy}{HTML}{043673}
\definecolor{rstint}{HTML}{EAF2FB}
\usepackage{titletoc}    
\usepackage{array}       

\input{header}  

\title{\ours: Unified Scraping and Cleaning \\ of Web Data for Effective LLM Pretraining}

\author{Zichun Yu\thanks{Equal contribution.}, Jiarui Yan$^{*}$, Shlok Sanghvi, Nihar Atri, Chenyan Xiong \\
Language Technologies Institute, Carnegie Mellon University \\
\texttt{\{zichunyu,cx\}@andrew.cmu.edu} \\[0.8em]
\normalfont
\lnk{\hficon}\textbf{Dataset}: \url{https://huggingface.co/datasets/cx-cmu/ReScraper-Data} \\
\lnk{\hficon}\textbf{Model}: \url{https://huggingface.co/cx-cmu/ReScraper} \\
\lnk{\raisebox{-0.1em}{\faGithub}}\textbf{Code}: \url{https://github.com/cxcscmu/ReScraper}
}

\iclrfinalcopy 
\begin{document}

\maketitle
\thispagestyle{fancy}
\fancyhead[L]{Preprint.}

\input{sections/abstract}
\input{sections/intro}
\input{sections/related}
\input{sections/method}
\input{sections/setup}
\input{sections/results}
\input{sections/conclusion}

\section*{Acknowledgments}
We thank Amazon for funding Zichun Yu through the Amazon AI Ph.D. Fellowship Program.
We thank Institute of Foundation Models (IFM) and CMU Foundation and Language Model (FLAME) Center
for providing support of computational resources.

\bibliography{bibliography}
\bibliographystyle{iclr2027_conference}

\appendix
\clearpage
\input{sections/appendix}

\end{document}

%% file: math_commands.tex
\usepackage{amsmath,amsfonts,bm}

\def\eqref#1{equation~\ref{#1}}

\def\1{\bm{1}}

\DeclareMathAlphabet{\mathsfit}{\encodingdefault}{\sfdefault}{m}{sl}
\SetMathAlphabet{\mathsfit}{bold}{\encodingdefault}{\sfdefault}{bx}{n}



%% file: header.tex
\definecolor{burntorange}{rgb}{0.8, 0.33, 0.0}

\usepackage{xcolor}
\usepackage{xspace}

\newcommand{\ours}{\textsc{ReScraper}\xspace}

\definecolor{midnightgreen}{rgb}{0.0, 0.29, 0.33}

\providecommand{\makecell}[1]{\shortstack{#1}}

\usepackage{hyperref}
\definecolor{codegreen}{rgb}{0,0.6,0}
\definecolor{prevcitegreen}{rgb}{0.0, 0.42, 0.24} %
\definecolor{citegreen}{rgb}{0.0, 0.42, 0.24} 
\definecolor{codegray}{rgb}{0.5,0.5,0.5}
\definecolor{codepurple}{rgb}{0.58,0,0.82}
\definecolor{backcolour}{rgb}{0.95,0.95,0.92}

\definecolor{midnightblue}{rgb}{0.11, 0.11, 0.6} %
\hypersetup{
  colorlinks=true, 
  linkcolor=midnightblue,  
  citecolor=midnightblue,   
}


%% file: sections/abstract.tex
\begin{abstract}
LLM pretraining corpora are normally cleaned by a stack of hand-written heuristics. A heuristic scraper
extracts the main content from HTML, and dozens of rule-based filters then clean it, so corpus
quality is capped by the coarseness and accuracy of the rules. In this work, we propose \ours, a unified
language model of only 0.6B parameters that replaces this entire stack. To train \ours, we carefully curate supervised data from the
outputs of three teacher models, so it learns to first extract the main content from raw data and
then choose among four operations: keeping the page as extracted, editing out noisy lines and spans,
deleting it entirely, or rewriting it when it is poorly written but informative.
Based on the same crawled data pool, pretraining 400M, 1.4B, and 2.8B models on our curated data improves
the DCLM Core score by a relative 3.8--4.7\% over the strongest baseline at each scale, including the costly multi-agent curation. Our analyses show that each operation plays a distinct and
complementary role, and that extracting and cleaning in one model outperforms a cascade of separate
models. \ours also concentrates its operations on the pages that need them, raising the quality of
poor pages the most while keeping the corpus diverse. These results demonstrate the feasibility and
effectiveness of AI4AI for pretraining data curation, where a small learned model takes over an
entire stage of the refining pipeline. 
\end{abstract}

%% file: sections/intro.tex
\section{Introduction}
\suppressfloats[t]  

\begin{wrapfigure}{r}{0.55\textwidth}
  \centering
  \vspace{-16pt}
  \begin{subfigure}[t]{0.228\textwidth}
    \centering
    \includegraphics[width=1.0\linewidth]{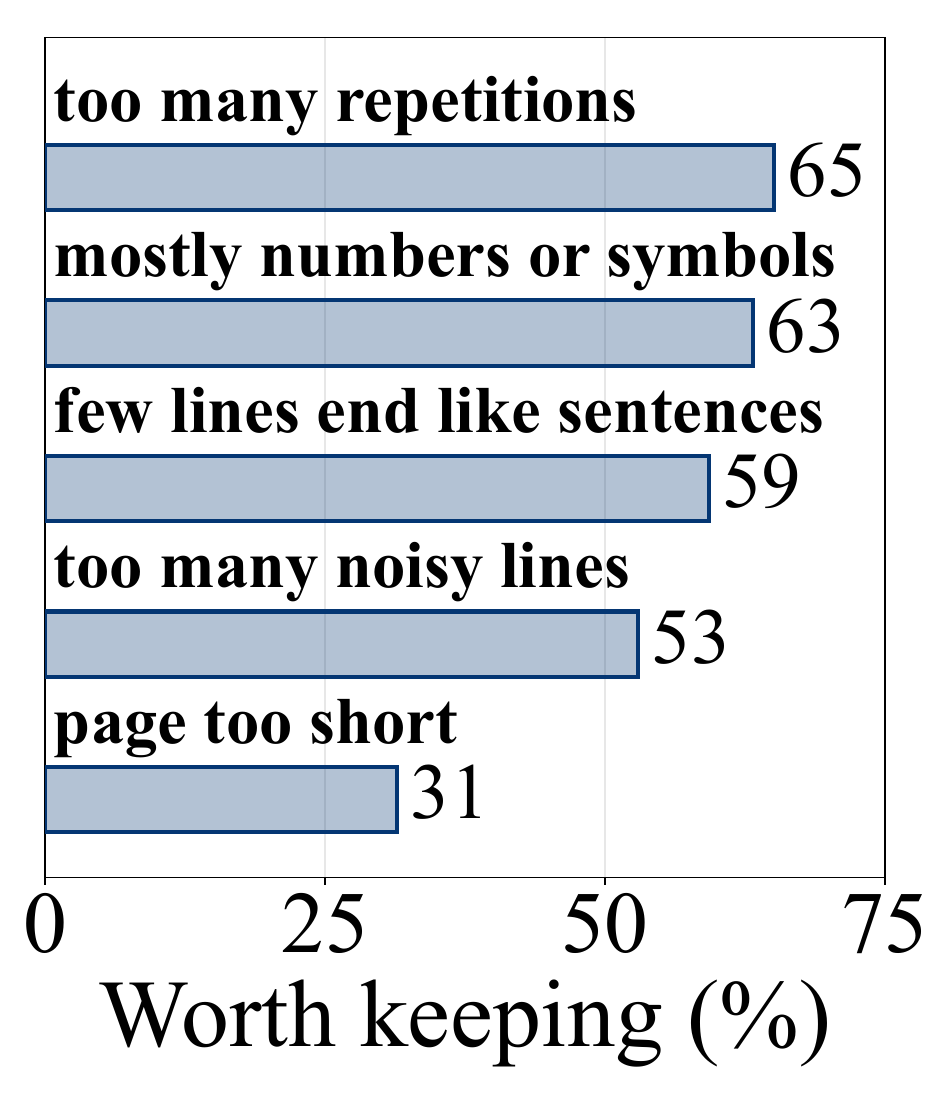}
    \caption{Rule drops}
    \label{fig:motivation-rules}
  \end{subfigure}
  ~
  \begin{subfigure}[t]{0.302\textwidth}
    \centering
    \includegraphics[width=1.0\linewidth]{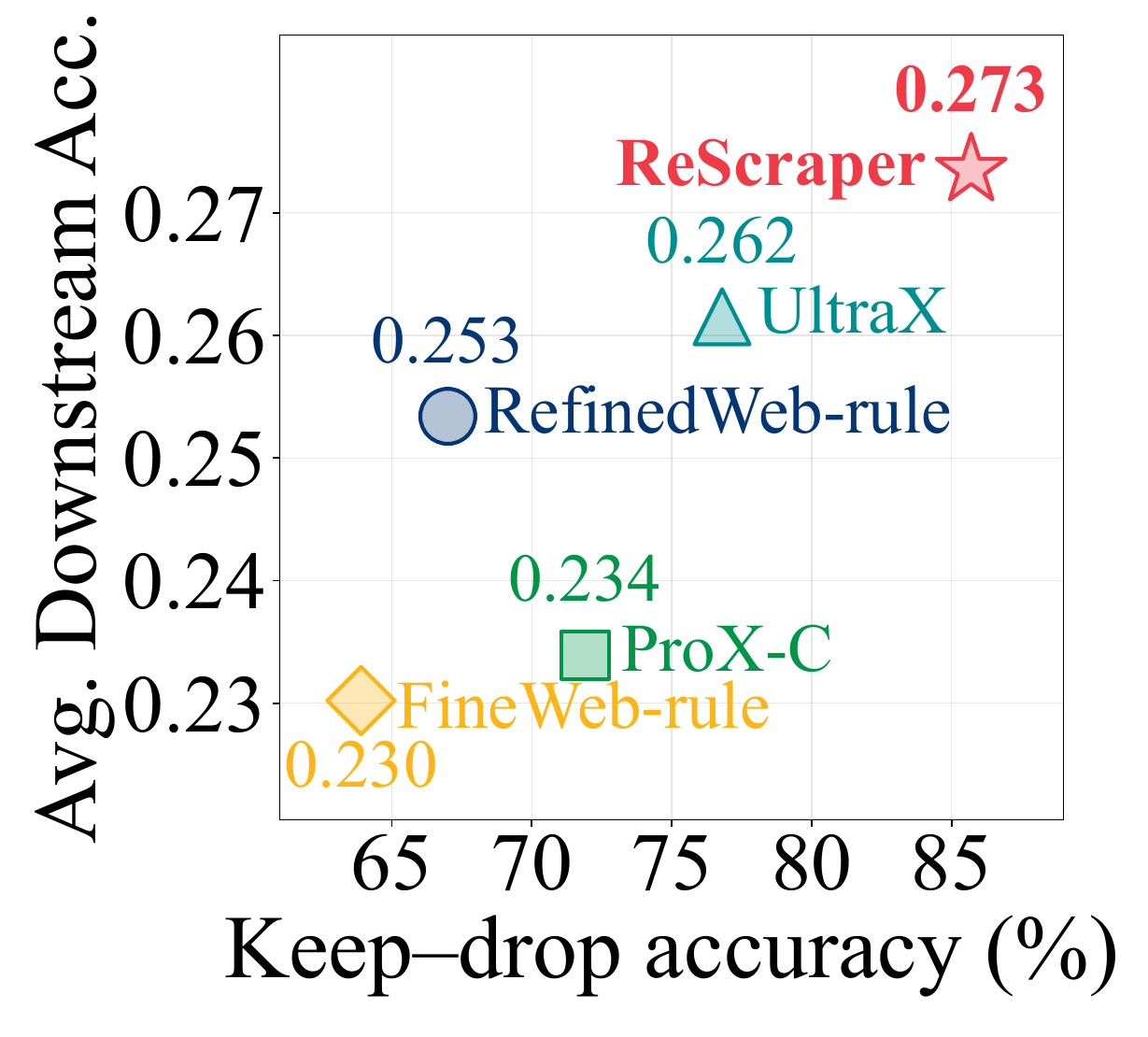}
    \caption{Keep--drop vs.\ downstream}
    \label{fig:motivation-keepdrop}
  \end{subfigure}
  \vspace{-0.25cm}
  \caption{Keep-or-drop decisions on 5,000 pages labelled by
    gpt-oss-120b~\citep{agarwal2025gpt}. (a) Share of the pages each rule drops that are
    worth keeping. (b) Keep--drop accuracy (agreement of each pipeline's keep or drop with the judge's label)
    against average downstream accuracy of 1.4B model pretraining.}
  \vspace{-0.2cm}
  \label{fig:motivation}
\end{wrapfigure}

Web crawls supply most LLM pretraining data~\citep{penedo2024fineweb,li2024datacomp}, and each
page must first be turned from raw data into clean, pretraining-ready text. This step still relies
on a classic heuristic pipeline, a rule-based scraper~\citep{bevendorff2018chatnoir,barbaresi2021trafilatura}
followed by filters on length, symbol ratios, and repetition~\citep{t5,penedo2024refinedweb,penedo2024fineweb}.
However, each rule mostly keeps or drops a whole page, so a page with one noisy paragraph either
loses its useful content or keeps its noise~\citep{zhou2024prox}. Furthermore, as shown in Figure~\ref{fig:motivation-rules}, the filter
rules could be inaccurate, with 31--65\% of the pages each rule drops judged worth keeping by an independent
LLM-as-a-judge~\citep{agarwal2025gpt}.

Learned models have begun to replace parts of this pipeline, including model-based
scrapers~\citep{xu2024neuscraper,liu2026dripper}, quality
filters~\citep{wettig2024qurating,peng2025dataman}, and language models that
edit~\citep{zhou2024prox,zhao2026ultrax} or rewrite~\citep{maini2024rephrasing,yu2025repro} web
text. Each still takes over one stage and consumes the output of the stage before it, so the
pipeline keeps a scraper whose mistakes no later model can undo, and errors compound from
one stage to the next. This raises a natural question. \textit{Can a single small language model
replace the entire pipeline, adaptively turning raw data into pretraining-ready text?}

In this paper, we introduce \ours, a unified and adaptive language model that takes raw data and
outputs pretraining-ready text. To handle the various operations a page may need, we train this
0.6B model on supervised data curated from the outputs of three teacher models, an extraction teacher, a
refining teacher, and a rewriting teacher. The data is constructed so that \ours learns to first \textit{extract}
the main content of each page and then choose to \textit{keep} it as extracted, \textit{edit} out
noisy lines and spans, \textit{delete} it, or \textit{rewrite} it to rescue educational content
that would otherwise be deleted. \ours thus replaces both the scraper and the cleaning pipelines,
and makes an adaptive decision for each page based on its content.

To evaluate the data \ours curates, we apply it to 18.0M Common Crawl
pages~\citep{li2024datacomp} and pretrain 400M, 1.4B, and 2.8B models on its output. Compared with
cascades that pair a scraper with widely adopted rule-based
cleaning~\citep{t5,penedo2024refinedweb,penedo2024fineweb} or with the best-performing model-based
refiners~\citep{zhou2024prox,zhao2026ultrax}, \ours achieves the highest DCLM
Core score (centered accuracy averaged over 22 downstream tasks) at every scale. Although \ours has only 0.6B parameters, it improves
1.4B and 2.8B pretraining by a relative 4.6\% and 3.8\% over the strongest baseline, enabling
weak-to-strong pretraining data curation. \ours even improves over the multi-agent data curation pipeline
DataOrchestra~\citep{huang2026dataorchestra} by a relative 6.6\% at 1.4B, showing the strong potential of a unified
small refiner.

To better understand where the gains come from, we first show through ablations that every operation of \ours contributes to the pretraining data,
and that performing extraction and cleaning together beats running them as separate stages.
Furthermore, \ours changes only the pages that need it: it improves poor pages the most, keeps good
ones, and does not make the corpus more repetitive. Finally, the 0.6B student closely follows its teachers
at a fraction of their cost. These results highlight the promise of building a unified model for
pretraining data curation.

Our contributions are summarized as follows.
\begin{enumerate}
  \item We propose \ours, a single small language model that replaces the
    whole heuristic scraping and cleaning pipeline, advancing AI4AI for pretraining data
    curation.
  \item We demonstrate weak-to-strong pretraining data curation, where data curated by the 0.6B \ours
    trains better 1.4B and 2.8B models than every baseline pipeline.
  \item We find that the unified model develops more effective refining behaviors,
    outperforming the best scraping+refining cascade with better efficiency.
\end{enumerate}

%% file: sections/related.tex
\section{Related work}
\vspace{-0.2cm}

\paragraph{HTML scraping.}
Turning a crawled page into text starts with a scraper that separates the main content
from navigation, advertisements, and templates. Heuristic scrapers such as
resiliparse~\citep{bevendorff2018chatnoir}, trafilatura~\citep{barbaresi2021trafilatura},
and jusText~\citep{pomikalek2011justext} decide with hand-written rules over the DOM tree
and link density, so they disagree across page layouts, and the union of several recovers
far more usable tokens than the best single one~\citep{li-etal-2026-beyond}. Recent work
instead trains a small language model to label HTML elements as content or
noise~\citep{xu2024neuscraper}, at trillion-token scale~\citep{ma2025aicc}.
Dripper~\citep{liu2026dripper} distills block-level annotations from a large LLM into a
0.6B model that marks each block of a simplified page as main content or not. The main purpose
of these scrapers is to identify the main content from the structure of the page.

\paragraph{Rule-based filtering and cleaning.}
Scraped web text still carries template text, non-informative pages, and duplicated
content. Pretraining data pipelines therefore apply document-level filters on
language~\citep{wenzek2020ccnet}, length and symbol statistics~\citep{rae2021scaling}, and
repetition~\citep{penedo2024refinedweb}, followed by
deduplication~\citep{broder1997resemblance,lee-etal-2022-deduplicating}. This recipe
underlies most widely used pretraining corpora, such as C4~\citep{t5}, Dolma~\citep{dolma}, FineWeb~\citep{penedo2024fineweb}, and
DCLM~\citep{li2024datacomp}. However, these
hand-crafted rules apply the same threshold to every page regardless of its content, which
makes them less accurate~\citep{zhou2024prox}. Additionally, they mostly decide to keep or drop a
whole page, so a page with one noisy paragraph is likely either discarded or kept whole.

\paragraph{Model-based filtering and refinement.}
As a fixed rule applies the same threshold to every page, model-based methods judge
quality more adaptively by learning it from the data. One family keeps the documents that
score highest under a fastText classifier~\citep{joulin2017bag,li2024datacomp},
reference-model perplexity~\citep{ankner2025perplexed}, an LLM
rating~\citep{sachdeva2024askllm,wettig2024qurating,peng2025dataman}, importance
resampling~\citep{xie2023data}, or data influence~\citep{yu2024mates}, but still keeps or
drops whole documents. Another
rewrites web pages with a language
model~\citep{maini2024rephrasing,su2025nemotroncc,nguyen2025rewire}, and
RePro~\citep{yu2025repro} trains the rewriter with a faithfulness reward to keep it anchored
to the source page. A third family has a model emit edit programs that are executed on the document instead of
regenerating it, with the edit space growing from deletion in
RefineX~\citep{bi2025refinex} to string replacement in ProX~\citep{zhou2024prox} and
insertion in UltraX~\citep{zhao2026ultrax}. DataOrchestra~\citep{huang2026dataorchestra}
goes a level higher with a multi-agent pipeline, in which a 1.7B orchestrator routes each
chunk to a 0.6B line-pruning model or a 4B rewriting model, and
DataEvolve~\citep{mi2026dataevolve} evolves cleaning strategies per data category. All of them consume text a
heuristic scraper has already extracted.
Our refiner instead starts from \textit{HTML} and performs all of these steps within a
\textit{single model}.

%% file: sections/method.tex
\section{Method}
\label{sec:method}

\begin{figure*}
  \centering
  \includegraphics[width=0.99\linewidth]{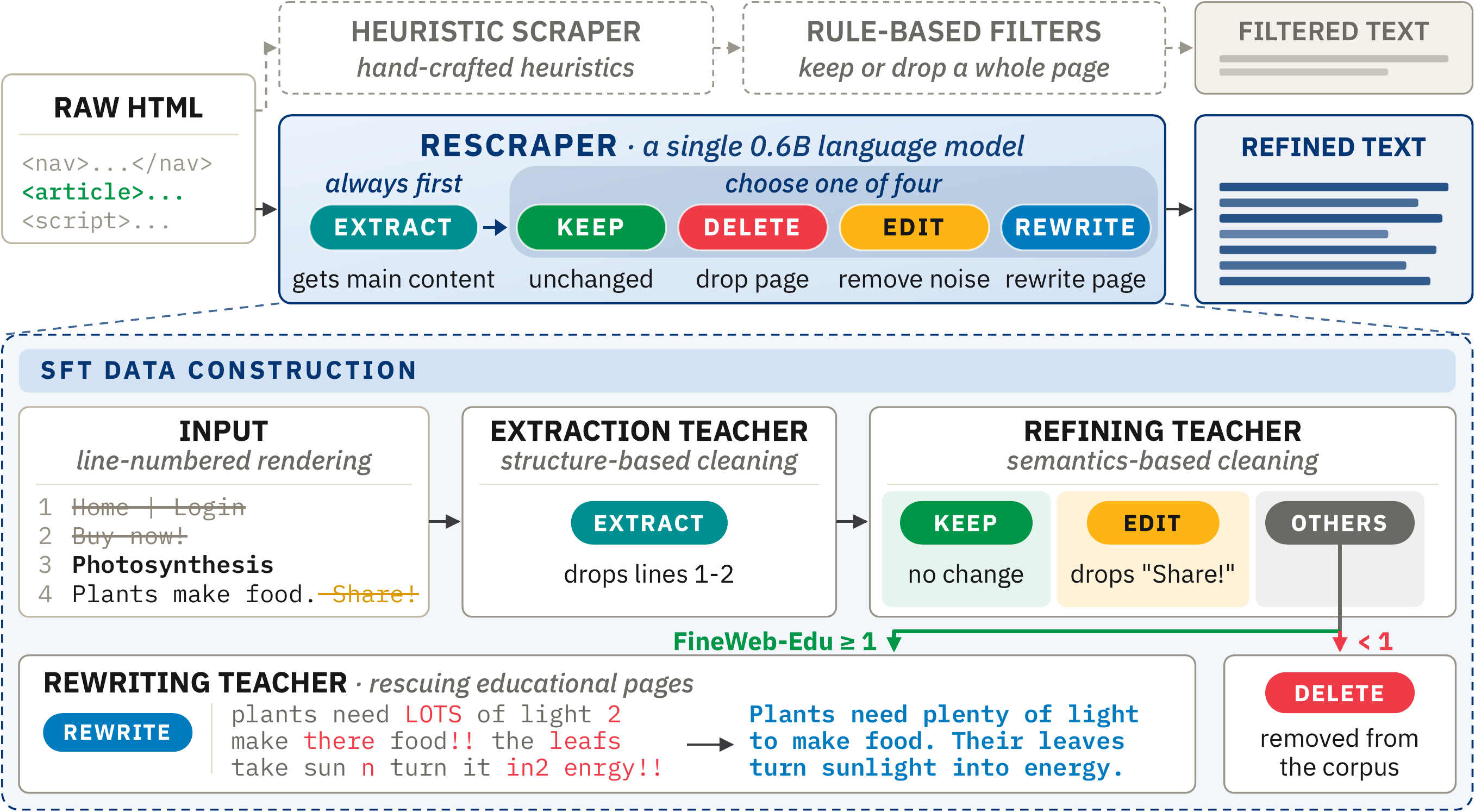}
  \caption{Overview of \ours. In place of a heuristic scraper followed by rule-based filters,
    a single small language model always extracts the main content first and then chooses one
    of four operations for the page. The bottom panel shows how three teachers label each page
    for SFT.}
  \label{fig:method}
  \vspace{-0.5cm}
\end{figure*}

In this section, we introduce \ours, our unified pretraining data curation model, as illustrated in Figure~\ref{fig:method}.
It replaces the heuristic scraper and rule-based filters with a single small language model
trained by a unified supervised fine-tuning. We first define the task and its operations
(\S\ref{sec:formulation}), then build the SFT targets from three teachers
(\S\ref{sec:sft-data}), and finally train \ours in two stages (\S\ref{sec:training}).

\subsection{Problem formulation}
\label{sec:formulation}

\ours turns raw data into clean pretraining text in a single generation. Its input keeps all the
visible text of a crawled page and changes only its format. Specifically, we render the raw HTML
to text, strip residual markup with BeautifulSoup, and
concatenate the text one block per line, prefixing each line with an identifier
\texttt{<lid:$n$>}. Navigation, sidebars, and footers thus remain in the input, and every removal
in the final text is made by the model.

For each page, \ours learns to generate a sequence of the form
\begin{tcolorbox}[colback=rstint, colframe=rsnavy, boxrule=0.5pt, arc=2pt, halign=center,
  left=2pt, right=2pt, top=2pt, bottom=2pt, before skip=4pt, after skip=4pt, fontupper=\small]
\mbox{\texttt{<extract>}~\textit{[extraction payload]}~~(\texttt{<keep>}\,$|$\,\texttt{<delete>}\,$|$\,\texttt{<edit>}\,$|$\,\texttt{<rewrite>})~\textit{[operation payload]}}
\end{tcolorbox}
\noindent where the extraction payload lists the
lines removed by \texttt{<extract>}, the tag in parentheses is the chosen operation, and the
operation payload specifies what that operation does, as shown in Table~\ref{tab:operations}. Extraction removes boilerplate, such as navigation and footers,
from every page without a decision, and the operation is then chosen from the content of the
extracted text. A deterministic executor applies this output to the rendered lines. After the
\texttt{<extract>} lines are dropped, \texttt{<keep>} returns the remaining text unchanged,
\texttt{<delete>} discards the page, \texttt{<edit>} removes the further lines and strings in
its payload, and \texttt{<rewrite>} replaces the page with its payload. Every operation except
\texttt{<rewrite>} only removes lines or strings addressed by line identifiers, so kept text respects original writing. This also keeps the output efficient, since it lists
only what to remove and its length grows with the number of removals rather than with the
length of the page.

\begin{table}[t]
  \centering
  \small
  \renewcommand{\arraystretch}{1.2}
  \setlength{\tabcolsep}{4pt}
  \caption{Operations of \ours. Every page first goes through \texttt{<extract>}, and the model
    then chooses one of the four operations below. Line numbers $a$ and $b$ are
    \texttt{<lid:$n$>} identifiers of the input.}
  \label{tab:operations}
  \begin{tabular}{>{\raggedright\arraybackslash}p{0.12\linewidth}>{\raggedright\arraybackslash}p{0.18\linewidth}>{\raggedright\arraybackslash}p{0.48\linewidth}l}
    \toprule
    \textbf{Operation} & \textbf{Payload} & \textbf{Explanation} & \textbf{Teacher} \\
    \midrule
    \texttt{<extract>} & \texttt{rm $a$-$b$}, \texttt{rm $a$} & Always performed first. Drops
      lines $a$ through $b$ (or line $a$) that are not main content. & Dripper \\
    \midrule
    \texttt{<keep>} & none & Return the extracted text unchanged. & Qwen3.8-27B \\
    \texttt{<delete>} & none & Discard the page from the pretraining corpus. & Qwen3.8-27B \\
    \texttt{<edit>} & \texttt{rm $a$-$b$}, \texttt{rm $a$} \newline \texttt{sub $a$: "$s$"} &
      Drop further noise lines $a$ through $b$ (or line $a$). \newline Delete the string $s$
      from line $a$. & Qwen3.8-27B \\
    \texttt{<rewrite>} & rewritten document & Replace the whole page with newly written text. &
      RePro \\
    \bottomrule
  \end{tabular}
\end{table}

\subsection{SFT data construction}
\label{sec:sft-data}

We build each SFT target by running three teachers in sequence on the same line-numbered
rendering, an extraction teacher, a refining teacher, and a rewriting teacher. Their outputs are
converted into operations over the same line identifiers and composed into a single target.

\paragraph{Main-content extraction (\texttt{<extract>}).}
To separate the main content of a page from boilerplate, advertisements, and navigation, we
adopt Dripper~\citep{liu2026dripper}, a lightweight language model trained for main-content
extraction from HTML, map its output back onto the rendered lines, and record the lines it
drops as \texttt{rm} operations, which form the payload of \texttt{<extract>}. Dripper does not judge whether
the main content is worth training on, and it never edits inside a line.

\paragraph{Semantic filtering and refinement (\texttt{<keep>}, \texttt{<delete>}, \texttt{<edit>}).}
To refine the extracted text at the semantic level, which structure-based extraction cannot
do, we adopt Qwen3.8-27B~\citep{qwen38}, a general-purpose language model that we prompt
with a fixed set of refinement rules (Appendix~\ref{sec:app-prompts}). A page with no training
value as a whole, such as spam, pure advertisements, or incoherent text, becomes
\texttt{<delete>} and is removed from the pretraining corpus unless the next step rescues it.
Every other page is cleaned only by deleting whole lines or fragments within a line, without
adding, changing, or reordering any word. If nothing is deleted, the extracted text is already
clean enough to train on and the page becomes \texttt{<keep>}. If anything is deleted, the page
becomes \texttt{<edit>}, which removes the noise that extraction leaves inside the main content. Its payload records the deleted
lines as \texttt{rm} operations and the deleted fragments as \texttt{sub} operations, as shown
in Table~\ref{tab:operations}.

\paragraph{Rescuing educational pages (\texttt{<rewrite>}).}
Some deleted pages still carry educational content, but in a form too noisy to train on that
removing lines or strings cannot repair, so \texttt{<edit>} does not apply, and deleting such a
page discards the information it contains. To identify these pages, we score each \texttt{<delete>} page with the
FineWeb-Edu classifier~\citep{penedo2024fineweb}, which rates the educational value of the
extracted text from 0 to 5. We rescue pages scoring at least 1.0, since a score of 1 denotes
some basic educational information even amid promotional content, and the rest remain
\texttt{<delete>}. To rewrite the rescued pages, we adopt RePro~\citep{yu2025repro}, a rephraser
trained to recycle web text for pretraining while staying faithful to its source, and use its
output, a coherent and fluent version of the page, as the payload of a \texttt{<rewrite>}
target. A rescued page is therefore relabeled \texttt{<rewrite>}
rather than \texttt{<delete>}, so \ours learns to choose among all four operations.

\subsection{Two-stage training}
\label{sec:training}

We fine-tune \ours on the serialized targets in two stages, so that it learns progressively from
easy to hard, first mainly the operations that only remove text and then
\texttt{<rewrite>}, which must generate a full page of new text.

\paragraph{Stage 1 (learning all operations).}
The first stage mostly teaches the model to identify the correct operation for each page and to
generate an accurate payload for every operation other than \texttt{<rewrite>}. These operations only
remove text, so their payloads are short lists of removals over line identifiers rather than a
full page of new text. To learn these easier
operations more effectively, we subsample \texttt{<rewrite>} to a small fraction of the
targets in this stage.

\paragraph{Stage 2 (strengthening rewrite).}
The second stage strengthens the learning of \texttt{<rewrite>}, which the first stage sees too
rarely to learn effectively. It continues from the first-stage checkpoint on a smaller mixture
that keeps every \texttt{<rewrite>} target and subsamples the other three operations, so the
model learns when a page should be rewritten rather than deleted and how to generate the
rewrite, while the other operations preserve what it learned in the first stage. The larger share
of \texttt{<rewrite>} targets also encourages the model to rewrite borderline pages instead of
deleting them.


%
%
%
%
%

%% file: sections/setup.tex
\section{Experimental setup}
\label{sec:setup}

\paragraph{Pretraining model and data.}
We pretrain decoder-only Transformers~\citep{vaswani2017attention} from scratch at three
scales, a 400M model on 8.2B tokens, a 1.4B model on 28.8B tokens, and a 2.8B model on
55.9B tokens. The three settings are the 400M-1x, 1B-1x, and 3B-1x scales of
DCLM~\citep{li2024datacomp}, which are set to be Chinchilla-optimal~\citep{Chinchilla}.
Every pipeline starts from the same source pool, 18.0M English Common Crawl documents
totaling 17.69B tokens, sampled i.i.d.\ from the DCLM pool.
The output of every pipeline is deduplicated with DCLM's Bloom filter.
The three budgets cover three regimes of data reuse. At 400M, the budget is close to the size of
each pipeline's output, the standard setting of about one epoch. At 1B, the output of \ours is
repeated about four times, within the range where repeated data is nearly as valuable as unique
data~\citep{muennighoff2023scaling}. At 3B, it is repeated more than four times, a data-bound regime
closer to the current bottleneck of scaling, as the stock of public human-generated text is
projected to run out~\citep{villalobos2024will}.

\paragraph{Baselines.}
We compare against three families of methods. All scrape with
resiliparse~\citep{bevendorff2018chatnoir}, the DCLM default.
\begin{enumerate}
  \item \textit{Rule-based cleaning}, the filter stacks of three representative corpora, C4~\citep{t5},
    RefinedWeb~\citep{penedo2024refinedweb}, and FineWeb~\citep{penedo2024fineweb}.
  \item \textit{Model-based refinement}, ProX-C~\citep{zhou2024prox} and
    UltraX~\citep{zhao2026ultrax}, which emit edit programs over the scraped text and so
    delete at the same granularities as \ours. For both we run the officially released refiner over our source pool.
  \item \textit{Multi-agent data curation}, DataOrchestra~\citep{huang2026dataorchestra}, in
    which a Qwen3-1.7B~\citep{yang2025qwen3} orchestrator decides for each chunk of the scraped text whether to drop,
    keep, or clean it, and sends chunks to be cleaned to a fine-tuned Qwen3-0.6B for line removal
    and to Qwen3-4B for instruction-guided rewriting.
\end{enumerate}

\paragraph{Evaluation.}
We evaluate the pretrained models on 22 downstream tasks from DCLM
Core~\citep{li2024datacomp} in either zero-shot or few-shot manners, covering commonsense
reasoning, language understanding, reading comprehension, symbolic problem solving, and
world knowledge. Our primary metric is centered accuracy, where per-task accuracy is
mapped to 0 for random guessing and 1 for perfect accuracy, and we report its average
across tasks as the Core score.

\paragraph{Implementation details.}
\ours is a Qwen3-0.6B model fine-tuned in two stages on targets built
as in \S\ref{sec:sft-data}, with 1.38M targets in the first stage and 131K in the second. Applying the student to the
source pool results in 7.44B unique tokens. The training hyperparameters, the composition of the targets, and
decoding settings are given in Appendix~\ref{sec:app-setup}, and all prompts in
Appendix~\ref{sec:app-prompts}.

%% file: sections/results.tex
\input{tables/html2text}

\section{Evaluation results}

In this section, we present main results (\S\ref{sec:results-main}), conduct ablations of the
operations of \ours and analyze its operation profile (\S\ref{sec:results-ablation}), the quality and
diversity of its output (\S\ref{sec:results-quality}), and how effectively it learns its teachers'
behaviors (\S\ref{sec:results-fidelity}).

\subsection{Main Results}
\label{sec:results-main}

\begin{figure}[t]
  \centering
  \includegraphics[width=0.9\linewidth]{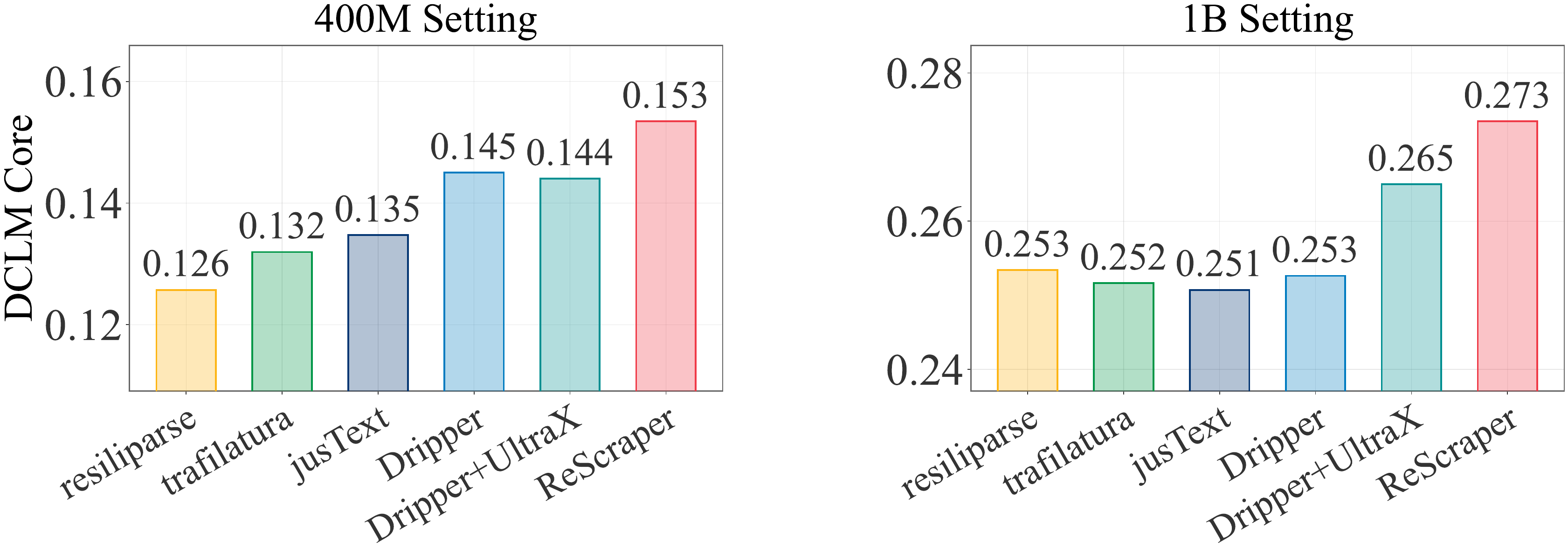}
  \vspace{-0.2cm}
  \caption{Core score in the 400M and 1B settings for each scraper (resiliparse,
    trafilatura, jusText, Dripper) followed by RefinedWeb-rule cleaning, the model-based
    baseline cascade (Dripper followed by UltraX), and \ours.}
  \label{fig:scraper-comparison}
\end{figure}

We first show the superiority of \ours in actual pretraining at all three scales. As shown in Table~\ref{tab:html2text}, \ours consistently outperforms all
baselines, improving over the strongest rule-based cleaning by a relative 4.7\%, 7.9\%, and 5.9\% at
400M, 1B, and 3B, while the strongest rule set itself changes with scale, from FineWeb-rule at 400M
to RefinedWeb-rule at 1B and 3B, indicating that no single set of hand-written rules is reliably
best. \ours, a 0.6B model, keeps its lead for the 1.4B and 2.8B models it curates data for, showing
weak-to-strong
pretraining data curation. Compared with the state-of-the-art multi-agent curation method
DataOrchestra, \ours improves Core by a relative 5.1\%, 6.6\%, and 3.8\% at 400M, 1B, and 3B, even though
DataOrchestra coordinates a 1.7B orchestrator with 0.6B and 4B tool models while
\ours is one 0.6B model. In the data-bound 3B setting, \ours repeats each unique token about 7.5
times, compared with 4.1 to 5.2 times for the model-based baselines, yet it
still outperforms them by a relative 3.8--7.2\%, indicating that its higher data quality outweighs
the cost of more repetition. These results show that one unified model, trained on
carefully curated supervision, curates more effective pretraining data than both the heuristic
stack and multi-model curation systems.

Furthermore, \ours outperforms the baseline pipelines regardless of which scraper they use. We apply RefinedWeb-rule
cleaning to the output of four scrapers, resiliparse~\citep{bevendorff2018chatnoir},
trafilatura~\citep{barbaresi2021trafilatura}, jusText~\citep{pomikalek2011justext}, and the
model-based Dripper~\citep{liu2026dripper}. As shown in Figure~\ref{fig:scraper-comparison}, \ours
outperforms all four at both scales, including its own extraction teacher Dripper, by a relative
5.8\% at 400M and 8.3\% at 1B. The choice of scraper also matters less as the pretrained model
grows, as the four scrapers span 0.019 Core at 400M but stay within 0.003 at 1B. Even the model-based baseline cascade, Dripper followed by UltraX, trails \ours by a relative 6.6\% at 400M and 3.2\%
at 1B, indicating that the advantage of \ours comes from extracting and cleaning each page
in one unified model, which no choice of scraper or cascade of separate stages can beat. \ours is
also more efficient, using about a third of the GPU hours of Dripper alone, as shown in
Table~\ref{tab:teacher-cost}.

\subsection{Effectiveness of Operations in \ours}
\label{sec:results-ablation}
\label{sec:results-operations}


\begin{table}[t]
  \centering
  \setlength{\tabcolsep}{1.5pt}
  \renewcommand{\arraystretch}{1.2}
  \caption{Ablation of the operations of \ours at the 1B scale. Dripper \texttt{<extract>}
    replaces the output of \texttt{<extract>} with Dripper's extracted text.}
  \label{tab:ablation-supervision}
  \resizebox{1.0\linewidth}{!}{%
  \begin{tabular}{l|c|cccccc}
    \toprule & \makecell{\\\textbf{\#Unique}} & \makecell{\textbf{Commonsense}\\\textbf{Reasoning}} & \makecell{\textbf{Language}\\\textbf{Understanding}} & \makecell{\textbf{Reading}\\\textbf{Comprehension}} & \makecell{\textbf{Symbolic}\\\textbf{Problem}} & \makecell{\textbf{World}\\\textbf{Knowledge}} & \makecell{\\\textbf{Core}} \\
    \textbf{Cleaning Method} &
    \textbf{Tokens} & \textit{(3 tasks)} & \textit{(6 tasks)} & \textit{(3 tasks)} & \textit{(5 tasks)} & \textit{(5 tasks)} & \textit{(22 tasks)} \\
    \midrule
    \rowcolor{blue!10} \ours & 7.44B & \textbf{0.33139} & \textbf{0.35533} & \textbf{0.20410} & \textbf{0.20563} & \textbf{0.24997} & \textbf{0.27348} \\
    ~~w/o \texttt{<edit>} & 7.77B & 0.30759 & 0.35021 & \underline{0.20224} & 0.20027 & \underline{0.24820} & \underline{0.26696} \\
    ~~w/o \texttt{<rewrite>} & 6.62B & 0.28903 & 0.34319 & 0.19418 & 0.16804 & 0.23994 & 0.25221 \\
    ~~w/o \texttt{<delete>} & 9.08B & 0.31067 & 0.33114 & 0.13187 & 0.18944 & 0.24550 & 0.24951 \\
    ~~w/o \texttt{<extract>} & 14.10B & 0.27084 & 0.29191 & 0.13419 & \underline{0.20447} & 0.20569 & 0.22806 \\
    \midrule
    Dripper \texttt{<extract>} & 7.34B & \underline{0.31803} & \underline{0.35208} & 0.19149 & 0.19647 & 0.23527 & 0.26363 \\
    \bottomrule
  \end{tabular}
  }
\end{table}

To examine how much each operation contributes to the pretraining data, we ablate each operation of
\ours in the 1B setting. Each variant disables a single operation when converting the model's
outputs into text. As shown in Table~\ref{tab:ablation-supervision}, removing any operation lowers
Core. Removing \texttt{<extract>} causes the largest drop, a relative 16.6\%, as boilerplate then
stays on every kept page, and removing \texttt{<delete>} costs 8.8\% and lowers every task
category, although both variants retain more unique tokens than
\ours. Removing \texttt{<rewrite>} lowers Core by 7.8\%, showing that rewriting recovers potentially
useful pages that would otherwise be deleted, and that these rewritten pages in turn help
pretraining. Removing \texttt{<edit>} has the smallest effect, a drop of 2.4\%, since it only
cleans lines and spans within pages that are kept anyway. Replacing the output of \texttt{<extract>} with Dripper's extracted text lowers Core by 3.6\%,
pointing to the advantage of performing extraction and cleaning in one unified model over a cascade of
separate models. These results indicate that every operation
contributes, and that accurate removal of low-quality content is essential.





\begin{figure}[t]
  \centering
  \begin{minipage}[t]{0.485\linewidth}
    \vspace{0pt}
    \centering
    \includegraphics[width=\linewidth]{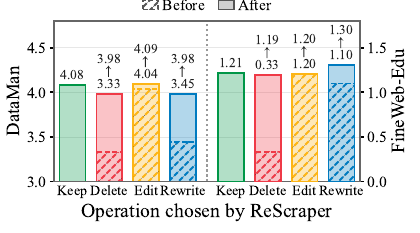}
    \caption{Mean quality scores of pages before and after each operation of \ours (for Delete, deleted
      pages vs.\ the rest).}
    \label{fig:operation-scores}
  \end{minipage}\hfill
  \begin{minipage}[t]{0.485\linewidth}
    \vspace{0pt}
    \centering
    \includegraphics[width=\linewidth]{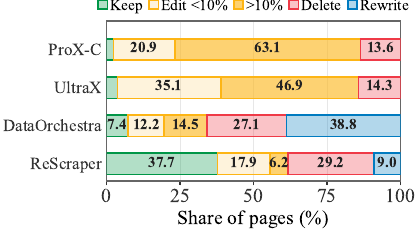}
    \caption{Share of pages per operation for each model-based pipeline; Edit is binned by the
      share of words removed.}
    \label{fig:operation-mix}
  \end{minipage}
\end{figure}

To understand
the underlying reasons for these gains, we score 5,000 pages held out from training with
DataMan~\citep{peng2025dataman} and the FineWeb-Edu classifier~\citep{penedo2024fineweb} before and
after the operation \ours chooses. As shown in
Figure~\ref{fig:operation-scores}, deleted pages score far below kept ones, editing slightly raises
DataMan without changing FineWeb-Edu, and rewriting raises both. Each operation thus addresses a
different problem, from low-quality pages to noise within a page and poorly written but informative
pages. As shown in Figure~\ref{fig:operation-mix}, \ours also changes pages sparingly, leaving 56\% of the pages
unchanged or with only small amounts of text removed, more than the other model-based refiners. \ours concentrates its changes on the pages that need them, with
operations that play distinct and complementary roles (further statistics in
Appendix~\ref{sec:app-additional-results}). The case studies in
Appendix~\ref{sec:app-cases} further show that \ours behaves more effectively than the cascaded pipelines, 
keeping structure that the scraper loses and deleting pages that the refiners keep.

\subsection{Quality and Diversity of Refined Data}
\label{sec:results-quality}

In this analysis, we compare the refined data at the level of individual pages and of the corpus
distribution, validating that \ours raises page quality while keeping the corpus diverse.

\begin{figure}[t]
  \centering
  \includegraphics[width=\linewidth]{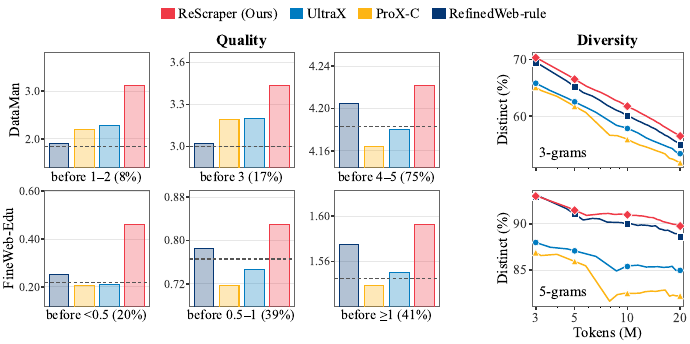}
  \vspace{-0.4cm}
  \caption{Left: mean quality score after cleaning on the held-out pages (share of pages in
    parentheses; dashed: mean before cleaning). Right: distinct $n$-gram share at equal token counts.}
  \label{fig:quality-diversity}
\end{figure}

\paragraph{\ours improves poor but informative pages and keeps good ones.} We group the held-out
pages by their score before cleaning and compare the mean score after each pipeline. As shown in
Figure~\ref{fig:quality-diversity} (left), \ours achieves the highest score in every group on both
scorers. The gap is largest on the poorest pages, where \ours raises DataMan by 1.28 compared with at
most 0.44 for the other pipelines, indicating that it is most effective where the source quality is
lowest. Meanwhile, it keeps 95\% of the pages that FineWeb-Edu rates 1.0 or higher, whereas
RefinedWeb-rule drops 30\% of them. This suggests that part of the
gain of \ours comes from turning low-quality web pages into useful training text rather than only
filtering them out.

\paragraph{\ours keeps the corpus diverse.} As illustrated in Figure~\ref{fig:operation-mix}, \ours deletes about twice as many
pages as ProX-C and UltraX, so we check that its corpus remains diverse. We
compare the share of distinct $n$-grams in each pipeline's output on the same source
pages, with every corpus cut to the same number of tokens. As shown in Figure~\ref{fig:quality-diversity} (right), \ours retains the highest share of distinct
3-grams and 5-grams,
confirming that its deletions do not make the corpus more repetitive, and thus it excels at both
quality and diversity.



\subsection{Effectiveness of Learning the Teachers' Behaviors}
\label{sec:results-fidelity}

In this analysis, we examine whether the 0.6B student preserves the behavior of its three teachers
(Dripper, Qwen3.8-27B and RePro), which determines whether it can replace them on new corpora.

\newlength{\sidefigheight}
\setlength{\sidefigheight}{0.274\textwidth}
\begin{figure}[t]
  \centering
  \begin{minipage}[t]{0.44\linewidth}
  \vspace{0pt}
  \centering
  \includegraphics[height=\sidefigheight]{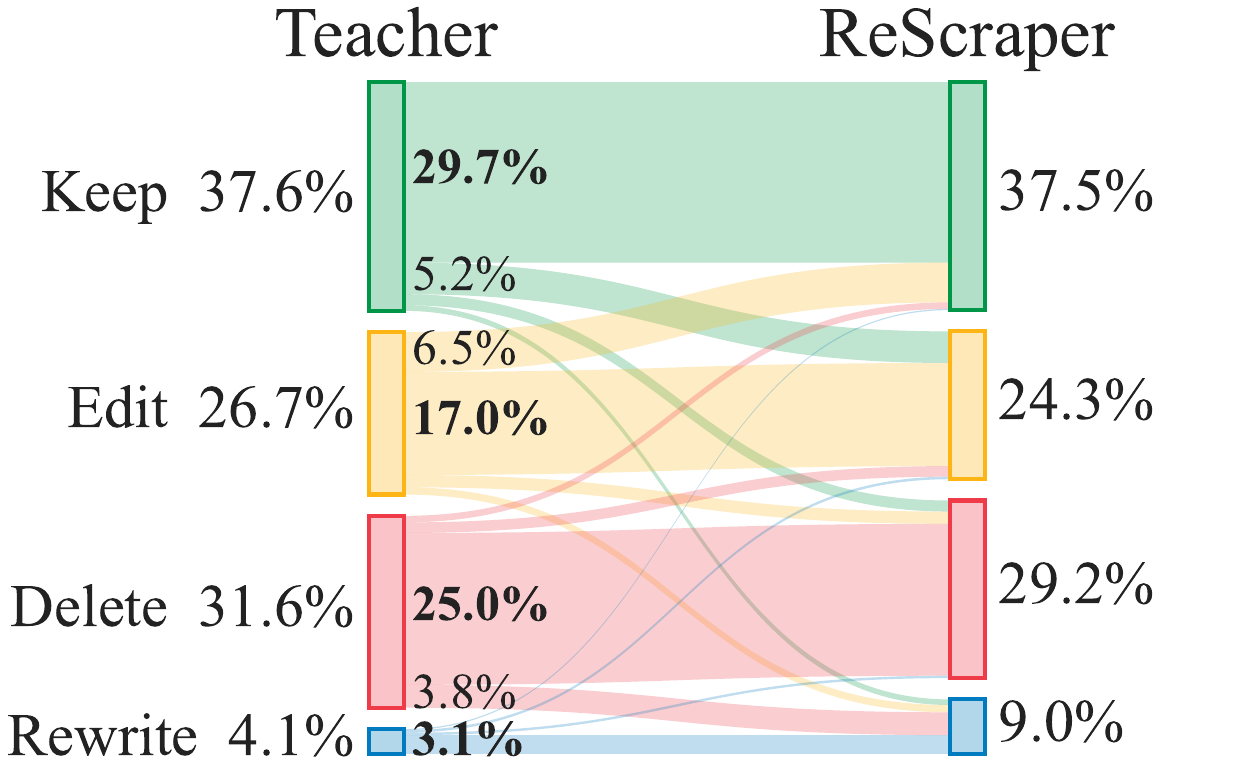}
  \caption{Operations chosen by the teacher and by \ours, in percent of pages.}
  \label{fig:decision-flow}
  \end{minipage}\hfill
  \begin{minipage}[t]{0.52\linewidth}
  \vspace{0pt}
  \centering
  \includegraphics[height=\sidefigheight]{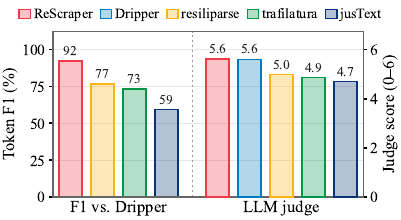}
  \caption{Token F1 against Dripper (left) and the gpt-oss-120b score without a reference
    (right).}
  \label{fig:extraction}
  \end{minipage}
  \vspace{-0.3cm}
\end{figure}

\paragraph{The student follows its teachers, keeps more tokens, and is cheap to run.} We label the
held-out pages with the training rules and compare the student's operations with the teacher's. As
shown in Figure~\ref{fig:decision-flow}, the student follows most of the teacher's operations, with a
similar distribution of operations over the pages, and reaches a token F1 of 89.3 against the
teacher's output. As demonstrated in Figure~\ref{fig:extraction}, its
extraction also matches Dripper's, and gpt-oss-120b~\citep{agarwal2025gpt} scores the two nearly
equally without a reference (prompt in Appendix~\ref{sec:app-prompt-extract-judge}). The main difference from the teacher is
intentional: in the second training stage, we increase the share of \texttt{<rewrite>} targets
(\S\ref{sec:training}) to encourage the student to rewrite borderline pages rather than delete them.
Accordingly, the student also rewrites some pages whose FineWeb-Edu score falls just below the
teacher's threshold of 1.0, retaining more tokens. Distillation thus keeps what the teachers do well and departs from them only
where we intended, at a fraction of their cost, as shown in Table~\ref{tab:teacher-cost}.

%% file: tables/html2text.tex
\begin{table*}
    [t]
    \centering
    \setlength{\tabcolsep}{1.5pt}
    \renewcommand{\arraystretch}{1.2}
    \caption{Benchmarking pretraining data curation methods at the 400M, 1B, and 3B scales. All baselines run on resiliparse-scraped text, while \ours curates the full rendered page directly.
    All scores are centered accuracies, mapped to 0 for random guessing and 1 for perfect accuracy, so negative values indicate below-random accuracy. $^*$ marks a multi-agent pipeline. \textbf{Bold} and \underline{underline} indicate the best and second-best results within each scale.}
    \vspace{-0.6cm}
    \label{tab:html2text}
    \vskip 0.15in
        \centering
        \resizebox{1.0\textwidth}{!}{%
        \begin{tabular}{l|c|cccccc}
            \toprule                                    & \makecell{\\\textbf{\#Unique}} & \makecell{\textbf{Commonsense}\\\textbf{Reasoning}} & \makecell{\textbf{Language}\\\textbf{Understanding}} & \makecell{\textbf{Reading}\\\textbf{Comprehension}} & \makecell{\textbf{Symbolic}\\\textbf{Problem}} & \makecell{\textbf{World}\\\textbf{Knowledge}} & \makecell{\\\textbf{Core}} \\
            \textbf{Cleaning Method} &
            \textbf{Tokens} & \textit{(3 tasks)} & \textit{(6 tasks)} & \textit{(3 tasks)} & \textit{(5 tasks)} & \textit{(5 tasks)} & \textit{(22 tasks)} \\
            \midrule
            \multicolumn{8}{l}{\textbf{400M Setting:} 400M model, 8.2B pretraining tokens} \\
            \midrule
            Raw text & 17.69B & 0.19822 & 0.14621 & \underline{0.03926} & 0.15624 & 0.11101 & 0.13300 \\
            C4-rule & 5.06B & 0.21326 & \underline{0.19160} & -0.01954 & 0.13270 & 0.12398 & 0.13701 \\
            RefinedWeb-rule & 7.04B & 0.18711 & 0.18266 & -0.07126 & 0.13612 & 0.12833 & 0.12572 \\
            FineWeb-rule & 5.16B & 0.20774 & \textbf{0.19588} & \textbf{0.06072} & 0.11123 & 0.13741 & \underline{0.14654} \\
            ProX-C & 11.80B & 0.19277 & 0.17242 & 0.03558 & 0.16270 & 0.12231 & 0.14294 \\
            UltraX & 10.78B & 0.17931 & 0.17712 & -0.02509 & \underline{0.17107} & 0.13747 & 0.13946 \\
            DataOrchestra$^*$ & 13.60B & \underline{0.21504} & 0.18097 & -0.04011 & 0.16702 & \textbf{0.15349} & 0.14605 \\
            \rowcolor{blue!10} \ours (Ours) & 7.44B & \textbf{0.22420} & 0.18741 & -0.01023 & \textbf{0.17347} & \underline{0.14843} & \textbf{0.15345} \\
            \midrule
            \multicolumn{8}{l}{\textbf{1B Setting:} 1.4B model, 28.8B pretraining tokens} \\
            \midrule
            Raw text & 17.69B & 0.28243 & 0.28718 & 0.15706 & 0.18036 & 0.20327 & 0.22544 \\
            C4-rule & 5.06B & 0.33107 & 0.32885 & 0.02322 & 0.12632 & 0.20640 & 0.21362 \\
            RefinedWeb-rule & 7.04B & 0.31254 & 0.34540 & 0.19665 & 0.16730 & 0.22765 & 0.25340 \\
            FineWeb-rule & 5.16B & \textbf{0.33412} & \underline{0.35349} & 0.08545 & 0.11129 & 0.22576 & 0.23022 \\
            ProX-C & 11.80B & 0.29379 & 0.31417 & 0.12276 & \underline{0.18323} & 0.21905 & 0.23391 \\
            UltraX & 10.78B & 0.30922 & 0.34396 & \textbf{0.21625} & 0.17274 & \underline{0.24991} & \underline{0.26152} \\
            DataOrchestra$^*$ & 13.60B & 0.30468 & 0.33381 & 0.19824 & 0.17711 & 0.24909 & 0.25648 \\
            \rowcolor{blue!10} \ours (Ours) & 7.44B & \underline{0.33139} & \textbf{0.35533} & \underline{0.20410} & \textbf{0.20563} & \textbf{0.24997} & \textbf{0.27348} \\
            \midrule
            \multicolumn{8}{l}{\textbf{3B Setting:} 2.8B model, 55.9B pretraining tokens} \\
            \midrule
            Raw text & 17.69B & 0.34258 & 0.37590 & 0.24421 & 0.16956 & 0.26102 & 0.28039 \\
            C4-rule & 5.06B & 0.31451 & 0.38711 & 0.12038 & 0.11294 & 0.22162 & 0.24091 \\
            RefinedWeb-rule & 7.04B & 0.36890 & 0.40926 & 0.22650 & 0.19516 & 0.27905 & 0.30058 \\
            FineWeb-rule & 5.16B & \underline{0.38285} & 0.40095 & 0.18665 & 0.13952 & 0.26515 & 0.27898 \\
            ProX-C & 11.80B & 0.35897 & 0.40405 & 0.21191 & 0.19486 & 0.28411 & 0.29690 \\
            UltraX & 10.78B & 0.37379 & \underline{0.42883} & 0.22526 & 0.18471 & \underline{0.28951} & 0.30642 \\
            DataOrchestra$^*$ & 13.60B & \textbf{0.38927} & 0.39583 & \textbf{0.24739} & \underline{0.20451} & 0.28857 & \underline{0.30683} \\
            \rowcolor{blue!10} \ours (Ours) & 7.44B & 0.38003 & \textbf{0.43042} & \underline{0.24443} & \textbf{0.20598} & \textbf{0.30387} & \textbf{0.31841} \\
            \bottomrule
        \end{tabular}
        }
    \vspace{-0.3cm}
\end{table*}

%% file: sections/conclusion.tex
\section{Conclusion}

In this paper, we introduce \ours, a unified and adaptive language model that turns raw web data
into pretraining-ready text. It replaces the heuristic scraping and cleaning stack with a single
0.6B model trained on the outputs of three teachers. The corpus it curates improves 2.8B pretraining
by a relative 3.8\% over the strongest baseline. This weak-to-strong pretraining data curation
suggests that the curator need not grow with the models it serves. A small model trained once can
prepare data for ever larger ones. Handling extraction and cleaning in one pass over the whole page
also lets the model adapt to each page in ways a cascade of separate models can hardly reproduce. We
hope \ours motivates future work toward curators that adapt the data they provide to the evolving
needs of the model being trained, and ultimately toward recursive self-improvement, where each
generation of models curates the pretraining data for the next.




%% file: sections/appendix.tex

\startcontents[app]
\printcontents[app]{l}{1}{
\section*{Appendix Table of Contents}}
\clearpage

\section{Detailed experimental setup}
\label{sec:app-setup}

The components of the \ours
pipeline and the pretraining hyperparameters~\citep{li2024datacomp} are given in Table~\ref{tab:impl-details}, the SFT
targets of each training stage in Table~\ref{tab:sft-composition}, and the cost of each step in
Table~\ref{tab:teacher-cost}. Both training stages upweight the loss on the operation tags by
$5\times$, so that a one-token decision is not outweighed by the removals and rewritten text that
follow it. The second training stage raises \texttt{<rewrite>} to 30\% of the mixture because its small
share in the first stage is not enough to learn the operation effectively. Each training stage uses its own
system prompt, and refinement of the pool uses the Stage 2 prompt, so the model is applied with the
output format it learned last. We run the Stage 2 model over the full rendered pages of the
source pool with the decoding settings in Table~\ref{tab:impl-details}, execute each output against
its line-numbered input as described in Section~\ref{sec:formulation}, and discard the outputs
rejected by the post-filter, which removes generations that leaked operation syntax into the text,
typically outputs truncated in the middle of an operation list. The retained text is deduplicated
with the DCLM Bloom filter and tokenized with the GPT-NeoX-20B tokenizer.


\input{tables/impl_details.tex}

\begin{table}[t]
  \centering
  \small
  \caption{SFT targets per operation in each training stage. Of the Stage 1 targets, 1,376,248 fit the
    32,768-token training sequence and are used for training.}
  \label{tab:sft-composition}
  \begin{tabular}{lrrrr}
    \toprule
    & \multicolumn{2}{c}{\textbf{Stage 1}} & \multicolumn{2}{c}{\textbf{Stage 2}} \\
    \cmidrule(lr){2-3}\cmidrule(lr){4-5}
    \textbf{Operation} & \textbf{Targets} & \textbf{Share} & \textbf{Targets} & \textbf{Share} \\
    \midrule
    \texttt{<keep>}    & 522,312   & 37.8\% & 35,189  & 26.8\% \\
    \texttt{<delete>}  & 499,924   & 36.1\% & 32,239  & 24.5\% \\
    \texttt{<edit>}    & 359,292   & 26.0\% & 24,611  & 18.7\% \\
    \texttt{<rewrite>} & 1,587     & 0.1\%  & 39,445  & 30.0\% \\
    \midrule
    Total              & 1,383,115 & 100\%  & 131,484 & 100\% \\
    \bottomrule
  \end{tabular}
\end{table}

\begin{table}[t]
  \centering
  \small
  \setlength{\tabcolsep}{3pt}
  \caption{Cost in H200 GPU hours from Slurm accounting, all attempts included. Teacher rows count the share of each teacher run spent on the pool pages the SFT set is drawn from. Stage 1 is counted up to the checkpoint stage 2 continues from.}
  \label{tab:teacher-cost}
  \begin{tabular}{llr}
    \toprule
    Paid & Stage & H200 GPU h \\
    \midrule
    Once       & Dripper labels              & 150 \\
    Once       & Qwen3.8-27B labels          & 204 \\
    Once       & RePro-1B rewrites ($T{=}1$) & 7 \\
    Once       & SFT, stage 1                & 244 \\
    Once       & SFT, stage 2                & 37 \\
    \midrule
    Per pool   & ProX-C        & 147 \\
    Per pool   & UltraX        & 180 \\
    Per pool   & \ours         & 494 \\
    Per pool   & Dripper       & 1,366 \\
    Per pool   & DataOrchestra & 1,652 \\
    \midrule
    Per corpus & 400M pretraining            & 26 \\
    Per corpus & 1.4B pretraining            & 240 \\
    Per corpus & 2.8B pretraining            & 740 \\
    \midrule
    \multicolumn{2}{l}{Total in this work} & 14,643 \\
    \bottomrule
  \end{tabular}
\end{table}

\section{Additional results}
\label{sec:app-additional-results}

In this section, we present the token retention of each stage of \ours (\S\ref{sec:app-retention}),
the selection of the final model (\S\ref{sec:app-ablation-extra}), the variance across pretraining
seeds (\S\ref{sec:app-seed-variance}), extended statistics of its operations
(\S\ref{sec:app-operation-stats}), the length of the documents each pipeline keeps (\S\ref{sec:app-token-dist}), the faithfulness of its
rewritten pages (\S\ref{sec:app-faithfulness}), and page quality under an LLM judge
(\S\ref{sec:app-quality-judge}).

\subsection{Token retention per stage}
\label{sec:app-retention}

\begin{wrapfigure}[10]{r}{0.28\textwidth}
  \centering
  \vspace{-32pt}
  \includegraphics[width=\linewidth]{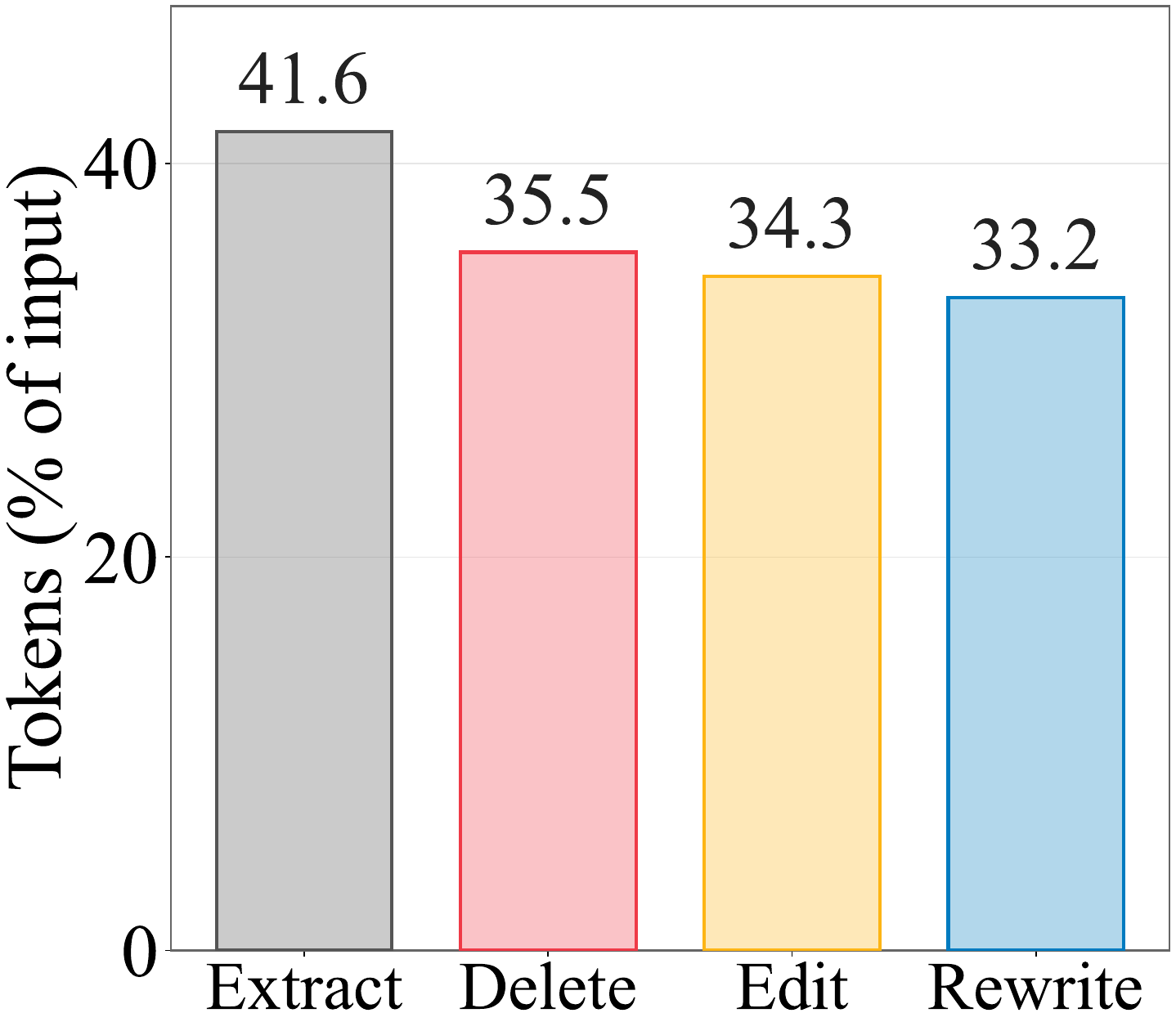}
  \caption{Share of input tokens left after each operation on the held-out pages.}
  \label{fig:token-waterfall}
  \vspace{-8pt}
\end{wrapfigure}

Figure~\ref{fig:token-waterfall} follows the tokens of the held-out pages through the model's
operations. Specifically, extraction removes 58\% of the tokens, deletion removes a further 6pp,
and editing and rewriting about 1pp each, since a rewritten page is on average shorter than
its extracted source.

\subsection{Selection of the final model}
\label{sec:app-ablation-extra}

Which deleted pages become \texttt{<rewrite>} targets, and how many of them the Stage~2 set
contains, determine how much text \texttt{<rewrite>} adds to the corpus. Before settling on the final
model, we therefore compared student variants with different data configurations.

\begin{figure}[t]
  \centering
  \includegraphics[width=0.9\linewidth]{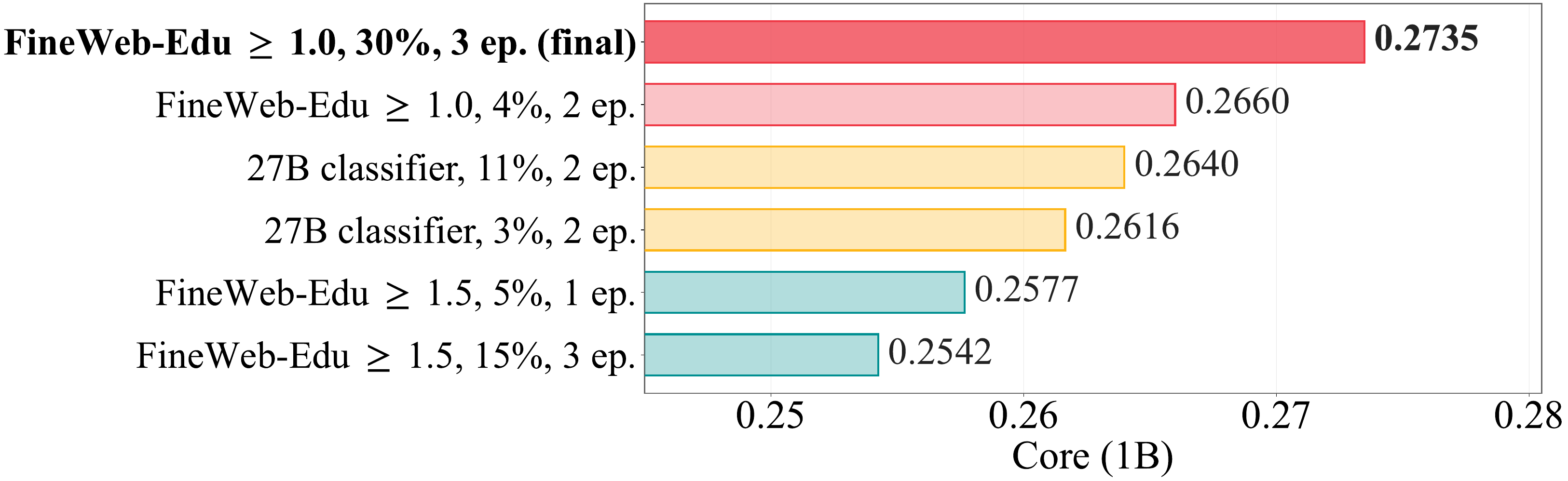}
  \caption{Core in the 1B setting of the corpora produced by student variants, labeled by
    rescue rule, share of \texttt{<rewrite>} in the Stage~2 set, and training epochs. All variants
    continue from the same Stage~1 checkpoint and are decoded like the final model.}
  \label{fig:ckpt-selection}
\end{figure}

All variants use RePro rewrites sampled at temperature 1.0 as
\texttt{<rewrite>} targets and are applied to the full pool with the decoding of
Table~\ref{tab:impl-details}; each corpus is post-filtered, deduplicated and tokenized like the
final one and used to pretrain a model in the 1B setting. They differ in the rule that selects
the rescued pages, in the share of \texttt{<rewrite>} in the Stage~2 set, and in training details such as the size of
that set, the number of epochs, and the system prompt. Besides the FineWeb-Edu
rule of \S\ref{sec:sft-data} at thresholds 1.0 and 1.5, we test a Qwen3.8-27B classifier that
judges whether a deleted page is worth rescuing. As shown in Figure~\ref{fig:ckpt-selection}, the
final student reaches the highest Core, 0.2735, and the other rules and mixes reach 0.2542 to
0.2660. Rescuing only pages
that score at least 1.5 gives at most 0.2577, and replacing FineWeb-Edu with the classifier gives
0.2616 to 0.2640.
Our final choice, the FineWeb-Edu rule at a threshold of 1.0 with \texttt{<rewrite>} making up 30\%
of the Stage~2 set, thus achieves the best result overall, and we use it throughout the paper.

\subsection{Variance across pretraining seeds}
\label{sec:app-seed-variance}

\begin{wrapfigure}[12]{r}{0.3\textwidth}
  \centering
  \vspace{-14pt}
  \includegraphics[width=\linewidth]{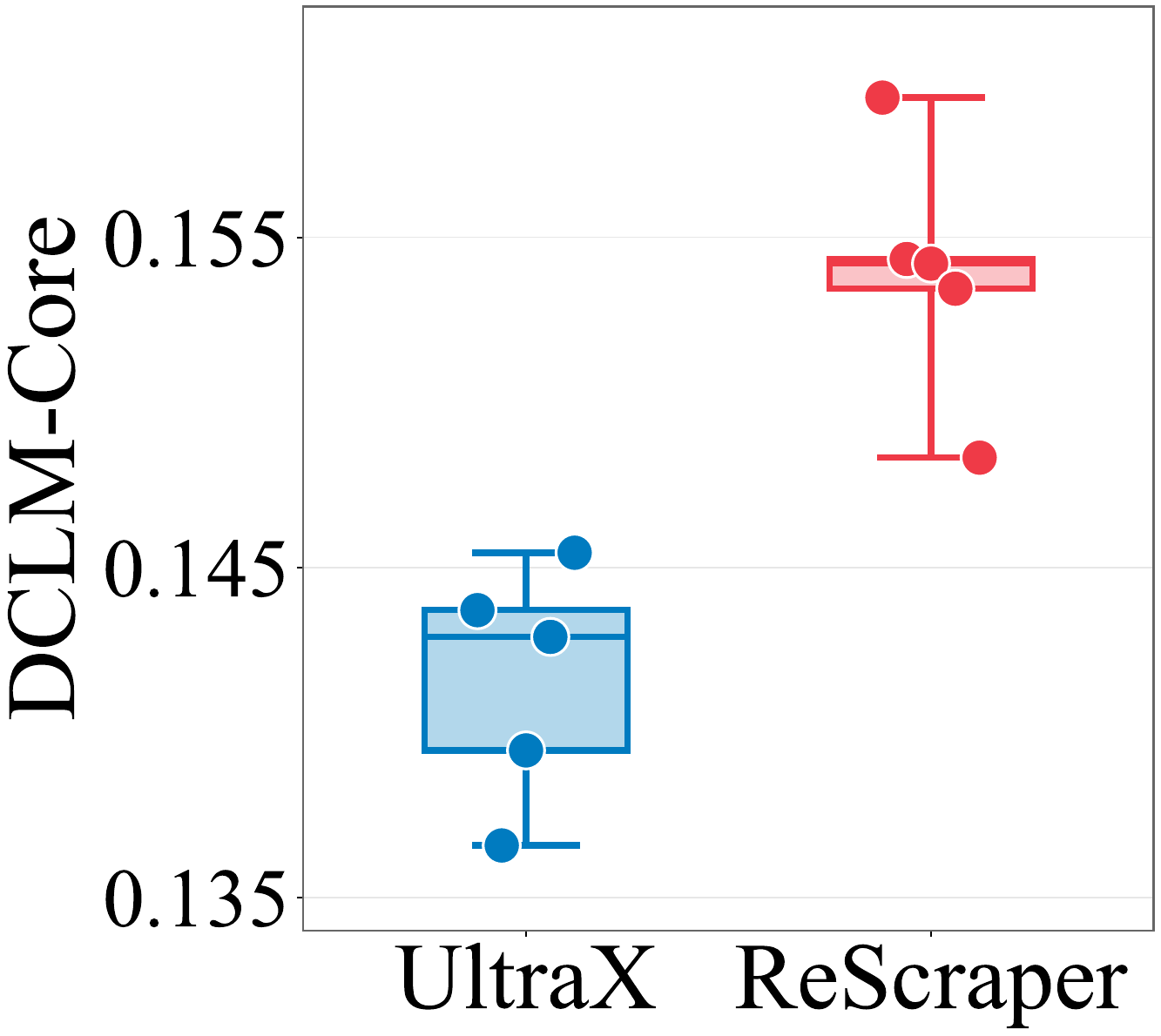}
  \vspace{-16pt}
  \caption{400M pretraining over 5 seeds of UltraX and \ours.}
  \label{fig:seed-variance}
  \vspace{-6pt}
\end{wrapfigure}

To estimate how much Core varies between pretraining runs, we pretrain the 400M model with five
random seeds on the corpus of \ours and on that of UltraX. Since pretraining is costly, we measure
the seed variance only at the smallest scale, following common practice~\citep{yu2025repro}. As shown
in Figure~\ref{fig:seed-variance},
Core averages 0.15392 for \ours with a standard deviation of 0.00387, and 0.14162 for UltraX with a
standard deviation of 0.00356; Table~\ref{tab:html2text} reports one run of each, 0.15345 and 0.13946.
The lowest seed of \ours, 0.14833, stays above the highest seed of UltraX, 0.14545, and the
difference between the two means is significant (Welch's $t$-test, $p<0.001$).

\subsection{Extended operation statistics}
\label{sec:app-operation-stats}

Table~\ref{tab:operation-examples} shows five examples of the outputs of \ours. In practice, we let
the model output its operation decision first, before the \texttt{<extract>} payload, since we find
it more effective to learn. Each output thus begins with the tag of the chosen operation, followed
by \texttt{<extract>} and its removals, the operation tag again, and the payload of the operation.
The executor reads the decision from the first tag and applies the \texttt{<extract>} removals to
every page whatever the decision, and the repeated tag marks where the payload begins.

\begin{table*}[t]
  \centering
  \scriptsize
  \setlength{\tabcolsep}{3pt}
  \caption{Five compact examples of model prediction and execution, drawn from the released
    SFT sample. The first target tag is the final decision; the \texttt{<extract>} block
    encodes Dripper's removals, and the repeated decision begins the Qwen/rewrite stage.
    A deterministic executor produces the corpus text and provenance \texttt{e2e\_tag}.
    Selected input lines are shown.}
  \label{tab:operation-examples}
  
  \begin{tabular}{p{0.08\textwidth}p{0.27\textwidth}p{0.28\textwidth}p{0.29\textwidth}}
    \toprule
    Case & Line-numbered input & Model prediction (SFT target) & Executed corpus record \\
    \midrule
    Dripper only & \texttt{<lid:1> Oswald passes out Fair Play for Cuba flyers on the streets of New Orleans, August 1963}
      & \texttt{<keep>} \newline \texttt{<extract>} \newline \texttt{<keep>}
      & \texttt{e2e\_tag=extract}; keep the Dripper-extracted line unchanged. \\
    Delete span & \texttt{<lid:1> IMG\_3785} \newline
      \texttt{<lid:2> Northrop Grumman's Vice President and Deputy General Counsel, Litigation and Investigations, Valerie Caproni (J.D.'79) shares her experiences with Georgia Law students / IMG\_3785 10/5/2012}
      & \texttt{<edit>} \newline \texttt{<extract>} \newline \texttt{rm 1} \newline
        \texttt{<edit>} \newline \texttt{sub 2: " / IMG\_3785 10/5/2012"}
      & \texttt{e2e\_tag=refine}; remove the image line and the exact trailing image identifier. \\
    Delete lines + span & \texttt{<lid:158> Supreme Court Scenarios ...} \newline
      \texttt{<lid:162> Supreme Court Scenarios ...} \newline
      \texttt{<lid:175> This short brief summarizes ... Also available: Detailed brief}
      & \texttt{<edit>} \newline \texttt{<extract>} \newline
        \texttt{rm 1--157; rm 159--161; rm 163;} \newline
        \texttt{rm 166--173; rm 176--196} \newline
        \texttt{<edit>} \newline \texttt{rm 158; rm 162} \newline
        \texttt{sub 175: " Also available: Detailed brief"}
      & \texttt{e2e\_tag=refine}; first apply Dripper's removals, then Qwen's two
        line removals and one exact-substring removal. \\
    Delete page & \texttt{<lid:1> The page has moved to: this page}
      & \texttt{<delete>} \newline \texttt{<extract>} \newline \texttt{<delete>}
      & \texttt{e2e\_tag=delete}; emit no document. \\
    Rewrite & \texttt{<lid:19> Web Design \& Devlopment} \newline
      \texttt{<lid:22> Website designing is an expansive term covering various aptitudes and controls that are utilized} \newline
      \texttt{<lid:23> as a part of the generation and upkeep of sites.}
      & \texttt{<rewrite>} \newline \texttt{<extract>} \newline
        \texttt{rm 1--18; rm 20--21; rm 42--63} \newline \texttt{<rewrite>} \newline
        \texttt{Web Design \& Development} \newline
        \texttt{Website designing encompasses a wide range of skills and tools used to create and maintain websites. ...}
      & \texttt{e2e\_tag=rewrite}; emit the newly generated replacement text. \\
    \bottomrule
  \end{tabular}
\end{table*}

\begin{table}[t]
  \centering
  \scriptsize
  \setlength{\tabcolsep}{3pt}
  \caption{Operations of \ours on the 4,989 held-out pages it processes, with the teacher's
    programs on the same pages for reference. An entry $m$ [$a$--$b$] ($\mu$) gives the median $m$,
    the interquartile range from $a$ to $b$, and the mean $\mu$.}
  \label{tab:operation-stats}
  \begin{tabular}{lrr}
  \toprule
  & \textbf{\ours} & \textbf{Teacher} \\
  \midrule
  \multicolumn{3}{l}{\texttt{<extract>} \textit{removals, all pages}} \\
  Pages & 4,989 & 4,989 \\
  Pages with a removal (\%) & 99.7 & 99.7 \\
  \texttt{rm} ops per page & 3 [2--4] (4.6) & 3 [2--4] (4.4) \\
  Lines removed per page & 85 [46--147] & 85 [46--146] \\
  Span length, lines & 6 [2--25] & 6 [2--27] \\
  Span length, words & 23 [7--107] & 25 [8--114] \\
  Removed lines: first / middle / last third (\%) & 34/31/35 & 34/31/35 \\
  Pages with a leading / trailing removed block (\%) & 96/98 & 96/98 \\
  Removed words: leading / interior / trailing (\%) & 31/12/58 & 30/11/59 \\
  \midrule
  \multicolumn{3}{l}{\texttt{<edit>} \textit{removals, edited pages}} \\
  Edited pages (ops / page text) & 1,134 / 82 & 1,212 / 120 \\
  \texttt{rm} ops per page & 1 [1--2] (1.9) & 1 [1--2] (1.8) \\
  \texttt{sub} ops per page & 0 [0--1] (0.8) & 0 [0--1] (0.8) \\
  \texttt{rm} only / \texttt{sub} only / both (\%) & 74/21/5 & 74/20/6 \\
  \texttt{rm} span, lines & 1 [1--2] & 1 [1--2] \\
  \texttt{rm} span, words & 6 [2--14] & 5 [2--13] \\
  \texttt{sub} span, words & 2 [1--3] & 2 [1--3] \\
  Edits: first / middle / last third (\%) & 29/30/40 & 30/29/40 \\
  Pages editing the first / last kept line (\%) & 23/51 & 24/53 \\
  \bottomrule
\end{tabular}
\end{table}

Table~\ref{tab:operation-stats} summarizes the operations \ours issues. \texttt{<extract>} removes
lines on 99.7\% of the pages, with a median of three line-removal operations covering 85 lines,
and these removals sit mostly at the page boundaries: 96\% of the pages lose a leading block and
98\% a trailing block, which together hold 89\% of the removed words. On edited pages, \ours
issues a median of one line removal and no substring removal, removing six words per line
removal or two words per substring removal, and 40\% of the edits fall in the last third of the
extracted text. These statistics closely match those of the teacher's programs on the same pages.

\subsection{Per-page token distributions}
\label{sec:app-token-dist}

Figure~\ref{fig:length-dist} compares document length across pipelines. The distributions
largely overlap; \ours and the two model-based refiners produce slightly shorter documents
than RefinedWeb-rule. On the 3,538 held-out pages \ours keeps, its text is shorter than the
resiliparse text of the same page on 93.5\% of the pages, with a median ratio of 0.75, since
extraction and editing remove text that resiliparse keeps. On the same pages, its documents are
close in length to those of the model-based refiners, with median ratios of 0.96 to UltraX and
0.98 to ProX-C. The rule-based pipelines instead cut the short end of the distribution: only
2.8\% of the documents of RefinedWeb-rule and 4.2\% of those of FineWeb-rule are under 100
tokens, against 10.5\% for \ours, since their length filters drop short pages whatever their
content, while \ours decides on each page by its content.

\subsection{Faithfulness of rewritten pages}
\label{sec:app-faithfulness}

\begin{figure}[!htb]
  \centering
  \begin{minipage}[t]{0.48\linewidth}
    \centering
    \includegraphics[width=\linewidth]{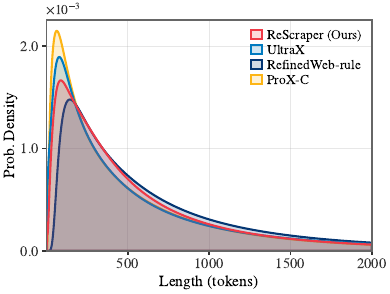}
    \caption{Length of the documents each pipeline keeps, in GPT-NeoX-20B tokens.}
    \label{fig:length-dist}
  \end{minipage}\hfill
  \begin{minipage}[t]{0.48\linewidth}
    \centering
    \includegraphics[width=\linewidth]{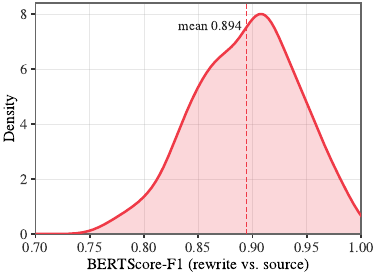}
    \caption{BERTScore-F1 between each rewritten held-out page and its extracted source.}
    \label{fig:rewrite-bertscore}
  \end{minipage}
\end{figure}

Since \texttt{<rewrite>} is the only operation that writes new text, we measure how closely each
rewritten page stays to its source. As shown in Figure~\ref{fig:rewrite-bertscore}, the rewrites reach a
mean BERTScore-F1 of 0.89 against the extracted text, indicating that \ours preserves the content
of the page while rephrasing it. The distribution peaks near 0.91, and almost all rewrites score
above 0.80, so the rewrites stay close to their sources. The rewrites are also slightly shorter
than their sources (\S\ref{sec:app-retention}), consistent with rephrasing that drops the
promotional and navigational fragments of a rescued page and keeps its informative content.

\subsection{Page quality under an LLM judge}
\label{sec:app-quality-judge}

Figure~\ref{fig:quality-diversity} (left) scores the text after cleaning with DataMan and the
FineWeb-Edu classifier, and FineWeb-Edu also selects the pages that \ours rescues. To check the
comparison with a scorer that plays no part in building \ours, we give every text that the pipelines
of Figure~\ref{fig:quality-diversity} keep from the held-out pages (15,106 texts) to
gpt-oss-120b~\citep{agarwal2025gpt} with the prompt in Appendix~\ref{sec:app-prompt-output-judge}.
The judge sees only the kept text, never the source page or a classifier score, and rates how much
informative content it holds from 0 to 3. We group the pages as in
Figure~\ref{fig:quality-diversity} and average the judge's rating over the pages each pipeline keeps.

\begin{table}[t]
  \centering
  \small
  \caption{Mean informative-content rating (0--3) of gpt-oss-120b for the texts each pipeline keeps
    from the held-out pages, grouped by the score of the page before cleaning as in
    Figure~\ref{fig:quality-diversity}.}
  \label{tab:quality-judge}
  \begin{tabular}{llrrrrr}
    \toprule
    \multicolumn{2}{l}{\textbf{Group before cleaning}} & \textbf{Pages} & \textbf{RefinedWeb-rule} &
      \textbf{ProX-C} & \textbf{UltraX} & \textbf{\ours} \\
    \midrule
    DataMan     & 1--2     &   375 & 0.59 & 0.56 & 0.71 & \textbf{1.09} \\
                & 3        &   852 & 0.88 & 0.91 & 1.02 & \textbf{1.32} \\
                & 4--5     & 3,772 & 1.69 & 1.66 & 1.73 & \textbf{1.90} \\
    \midrule
    FineWeb-Edu & $<$0.5   & 1,015 & 0.67 & 0.76 & 0.81 & \textbf{1.26} \\
                & 0.5--1   & 1,926 & 1.37 & 1.31 & 1.44 & \textbf{1.61} \\
                & $\geq$1  & 2,059 & 1.92 & 1.87 & 1.93 & \textbf{1.98} \\
    \bottomrule
  \end{tabular}
\end{table}

As shown in Table~\ref{tab:quality-judge}, \ours receives the highest rating in every group, and
the gap is again largest on the poorest pages: on pages that DataMan rates 1 or 2, the judge gives
the texts of \ours 1.09 against at most 0.71 for the other pipelines, and on pages that FineWeb-Edu
scores below 0.5, 1.26 against at most 0.81. The judge also finds a larger share of the texts of
\ours worth keeping in every group, for example 51\% against at most 32\% on the DataMan 1--2 pages.
The ranking of Figure~\ref{fig:quality-diversity} therefore does not depend on the scorer that
defines the rescue rule.

\clearpage  
\input{sections/cases.tex}

\clearpage  
\input{sections/prompts.tex}

%% file: tables/impl_details.tex
\begin{table}[t]
  \centering
  \renewcommand{\arraystretch}{1.3}
  \setlength{\tabcolsep}{3pt}
  \resizebox{0.95\linewidth}{!}{%
    \begin{tabular}{lr}
      \toprule
      \textbf{Component} & \textbf{Setting} \\
      \hline
      \multicolumn{2}{l}{\textit{Model and input}} \\
      Refiner backbone & Qwen3-0.6B~\citep{yang2025qwen3} \\
      Input rendering & HTML-to-text rendering, BeautifulSoup markup strip, \texttt{<lid:$n$>} ids \\
      Context window & 32,768 tokens (prompt $\leq$ 29,696, answer $\leq$ 3,072) \\
      Omitted pages & prompt over 29,696 tokens, rendering empty or under 20 characters, \\
       & or rendering over 60\,s \\
      \hline
      \multicolumn{2}{l}{\textit{SFT labels}} \\
      \texttt{<extract>} teacher & Dripper~\citep{liu2026dripper} \\
      \texttt{<keep>} / \texttt{<edit>} / \texttt{<delete>} teacher & Qwen3.8-27B~\citep{qwen38} \\
      \texttt{<rewrite>} teacher & RePro 1B rephraser~\citep{yu2025repro} \\
      Rewrite decoding & temperature 1.0, top-$p$ 0.9 \\
      Rescue classifier & FineWeb-Edu~\citep{penedo2024fineweb} \\
      Rescue threshold & score $\geq$ 1.0 \\
      \hline
      \multicolumn{2}{l}{\textit{SFT training}} \\
      Stage 1 epochs / steps / batch size & 3 / 3,366 / 192 \\
      Stage 2 epochs / steps / batch size & 3 / 336 / 192 \\
      Stage 2 peak learning rate & 8e-5 \\
      Learning-rate schedule & linear warmup ($\approx$11 steps), cosine decay \\
      Max sequence length & 32,768 tokens, packed \\
      Precision & bf16 \\
      Operation-tag loss weight (both stages) & 5$\times$ \\
      \hline
      \multicolumn{2}{l}{\textit{Inference and corpus construction}} \\
      Inference framework & vLLM 0.11.1 \\
      Decoding temperature / top-$p$ & 1.0 / 1.0 \\
      Max new tokens & 3,072 \\
      Thinking mode & disabled \\
      Post-filter & drop outputs with leaked operation syntax (0.061\%) \\
      Deduplication & DCLM Bloom filter, 13-grams, threshold 0.8 \\
      Tokenizer & GPT-NeoX-20B \\
      \hline
      \multicolumn{2}{l}{\textit{Pretraining}} \\
      Steps (400M / 1.4B / 2.8B) & 7,820 / 54,923 / 106,604 \\
      Batch size (400M / 1.4B / 2.8B) & 512 / 256 / 256 \\
      Sequence length & 2,048 \\
      Optimizer and schedule & AdamW, max learning rate 3e-3, cosine decay \\
      \bottomrule
    \end{tabular}
  }
  \caption{Implementation details of \ours and of the pretraining runs.}
  \label{tab:impl-details}
\end{table}

%% file: sections/cases.tex
\providecommand{\lid}[1]{\texttt{<lid:#1>}}
\providecommand{\cut}{\textit{[\ldots]}}
\providecommand{\op}[1]{\texttt{#1}}

\section{Case studies}
\label{sec:app-cases}

We show held-out pages of \S\ref{sec:results-operations} on which the model-based baselines fail
(Table~\ref{tab:case-baselines}), on which resiliparse loses structure (Table~\ref{tab:case-structure}),
and on which \ours applies its operations (Table~\ref{tab:case-operations}). ProX-C
and UltraX edit the resiliparse text of each page; \ours reads the full rendered page with line
identifiers (\S\ref{sec:formulation}), and its outputs are single samples under the decoding of
Table~\ref{tab:impl-details}. Text is verbatim except where it is cut at \cut; descriptions in
italic parentheses are ours.

\begin{table*}[t]
  \centering
  \scriptsize
  \setlength{\tabcolsep}{3pt}
  \caption{Failures of the model-based baselines. ProX-C and UltraX edit the resiliparse text of
    the page (second column), while \ours reads the full rendered page. The operations \ours
    generates are shown above its output.}
  \label{tab:case-baselines}
  \begin{tabular}{>{\raggedright\arraybackslash}p{0.08\textwidth}>{\raggedright\arraybackslash}p{0.24\textwidth}>{\raggedright\arraybackslash}p{0.12\textwidth}>{\raggedright\arraybackslash}p{0.22\textwidth}>{\raggedright\arraybackslash}p{0.26\textwidth}}
    \toprule
    Case & Resiliparse text & ProX-C & UltraX & \ours \\
    \midrule
    (a) Over-deletion and missed navigation
      & Presidio County, Texas \newline From Ballotpedia \newline Jump to: navigation, search \cut
        \newline Presidio County, Texas is one of 254 counties in Texas. \newline Website evaluation
        \cut \newline Local ballot measures in Texas \newline A-B \newline
        Anderson\allowbreak Andrews\allowbreak Angelina\allowbreak Aransas\cut \newline \cut
      & \textit{(empty)} \newline \op{remove\_lines(}\allowbreak\op{start=0, end=74)} \newline
        \op{remove\_lines(}\allowbreak\op{start=75, end=77)}
      & Presidio County, Texas \newline Presidio County, Texas is one of 254 counties in Texas.
        \newline Website evaluation \cut \newline Local ballot measures in Texas \newline A-B \newline
        Anderson\allowbreak Andrews\allowbreak Angelina\allowbreak Aransas\cut \newline \cut
      & \op{<keep> <extract> rm 1; rm 3-11; rm 22-94 <keep>} \newline
        Presidio County, Texas \newline Presidio County, Texas is one of 254 counties in Texas.
        \newline Website evaluation \newline The good \newline - The names of all county
        commissioners are provided. [1] \cut \newline The bad \newline - Individual contact
        information is not provided for the county commissioners. \cut \\
    \midrule
    (b) Valueless page kept
      & MySQL Error! \newline MySQL error in file: /engine/classes/mysql.php at line 52 \newline
        Error Number: 1 \newline The Error returned was: \newline Too many connections \newline
        SQL query:
      & \textit{(unchanged)} \newline \op{untouch\_doc()}
      & \textit{(unchanged)}
      & \op{<delete> <extract> rm 1 <delete>} \newline \textit{(no document)} \\
    \midrule
    (c) Sidebar scraped in place of the article
      & Trending \newline Trending Stories \newline Coconut oil may be worse than palm oil for the
        environment \newline \cut\ \textit{(four more headlines)} \newline Follow Us \cut
      & Coconut oil may be worse than palm oil for the environment \newline \cut\ \textit{(five
        headlines)} \newline \op{remove\_lines(}\allowbreak\op{start=0, end=3)} \cut
      & Coconut oil may be worse than palm oil for the environment \newline \cut\ \textit{(five
        headlines)} \newline Audra McDonald turns 40: a look back
      & \op{<keep> <extract> rm 1-30; rm 34; rm 38-56 <keep>} \newline Climate change threatens
        Africa water \newline TSHWANE, South Africa, Nov. 14 (UPI) -- Climate change in Africa's
        river basins could hamper the continent's farm transformation efforts \cut\ \textit{(the
        full article)} \\
    \bottomrule
  \end{tabular}
\end{table*}

\paragraph{Baseline failures.}
In Table~\ref{tab:case-baselines}(a), ProX-C deletes a county page, and UltraX keeps its article
together with the site's navigation boxes, while \ours keeps only the article. In
(b), a database error page, ProX-C and UltraX both return it unchanged, while \ours deletes it.
In (c), resiliparse returns a list of trending headlines instead of the news article, so both
baselines keep unrelated headlines; \ours reads the full rendered page and keeps the article.

\begin{table*}[t]
  \centering
  \scriptsize
  \setlength{\tabcolsep}{3pt}
  \caption{Structure lost in the resiliparse extraction. ProX-C and UltraX receive the text in the
    second column; \ours reads the full rendered page, in which the structure is intact.}
  \label{tab:case-structure}
  \begin{tabular}{>{\raggedright\arraybackslash}p{0.08\textwidth}>{\raggedright\arraybackslash}p{0.27\textwidth}>{\raggedright\arraybackslash}p{0.29\textwidth}>{\raggedright\arraybackslash}p{0.29\textwidth}}
    \toprule
    Case & Resiliparse text & \ours input (selected lines) & \ours \\
    \midrule
    (a) Table cells fused
      & Summary data for Cessna Citation Sovereign - \newline Country of originFirst
        flightNo. builtNo. in serviceCrewPassengers \newline USA200222522528 - 12 \cut \newline
        Dimensions for Cessna Citation Sovereign - \newline Length63 ft. 6 in. \newline Wing
        Span63 ft. 1 in. \newline Height20 ft. 1 in. \cut \newline Performance of Cessna
        Citation Sovereign - \newline Max Speed458 kts \newline Cruise Speed437 kts \cut
      & \lid{43} Country of origin First flight No. built No. in service Crew Passengers
        \newline \lid{44} USA 2002 225 225 2 8 - 12 \cut \newline \lid{49} Length 63 ft. 6 in.
        \newline \lid{50} Wing Span 63 ft. 1 in. \newline \lid{51} Height 20 ft. 1 in. \cut
        \newline \lid{55} Max Speed 458 kts \newline \lid{56} Cruise Speed 437 kts \cut
      & \op{<keep> <extract> rm 1-40; rm 60-95; rm 113-128 <keep>} \newline Aircraft Data
        \newline The Cessna Citation Model 680 Sovereign is a high performance, twin engined
        medium range business jet. \cut \newline Country of origin First flight No. built No.
        in service Crew Passengers \newline USA 2002 225 225 2 8 - 12 \cut \newline Length 63
        ft. 6 in. \newline Wing Span 63 ft. 1 in. \newline Height 20 ft. 1 in. \cut \\
    \midrule
    (b) List dropped
      & 2.18 Postinstallation Setup and Testing \newline Prev Chapter 2 Installing and
        Upgrading MySQL Next \cut \newline After installing MySQL, there are some items that
        you should address. For example: \newline Prev Up Next \newline 2.17.5 Compiling and
        Linking an Optimized mysqld Server Home 2.18.1 Unix Postinstallation Procedures
      & \lid{7} After installing MySQL, there are some items that you should address. For
        example: \newline \lid{8} - You should initialize the data directory and create the
        MySQL grant tables, as describe in Section 2.18.1, “Unix Postinstallation Procedures” .
        \newline \lid{9} - An important security concern is that the initial accounts in the
        grant tables have no passwords. \cut \newline \lid{10}--\lid{12} \textit{(three more
        items)} \newline \lid{13} Prev Up Next
      & \op{<keep> <extract> rm 1-2; rm 13-14 <keep>} \newline 2.18 Postinstallation Setup and
        Testing \cut \newline After installing MySQL, there are some items that you should
        address. For example: \newline - You should initialize the data directory and create
        the MySQL grant tables, \cut \newline - An important security concern is that the
        initial accounts in the grant tables have no passwords. \cut \newline \textit{(three
        more items)} \\
    \bottomrule
  \end{tabular}
\end{table*}

\paragraph{Structure recovery.}
In Table~\ref{tab:case-structure}(a), resiliparse fuses the cells of a specification table
(``225 225'' becomes ``225225''), and in (b) it drops the bulleted list that forms the body of a
documentation page. The baselines receive this text and cannot restore it, while the rendering \ours
reads keeps each table row and list item on its own line, and \ours keeps them.

\begin{table*}[t]
  \centering
  \scriptsize
  \setlength{\tabcolsep}{3pt}
  \caption{Operations of \ours on two held-out pages: the program it generates and the text the
    executor produces from it. Selected input lines are shown.}
  \label{tab:case-operations}
  \begin{tabular}{>{\raggedright\arraybackslash}p{0.08\textwidth}>{\raggedright\arraybackslash}p{0.33\textwidth}>{\raggedright\arraybackslash}p{0.29\textwidth}>{\raggedright\arraybackslash}p{0.23\textwidth}}
    \toprule
    Case & Line-numbered input & \ours program & Executed text \\
    \midrule
    (a) \op{<edit>}
      & \lid{1} Moncton Home Price Index for October 2015 \newline \lid{2} Moncton, Dieppe,
        Riverview, Shediac: Real Estate market news, stats, updates, 1st time homebuyer tips and
        more .... \newline \lid{3}--\lid{8} \textit{(site menu)} \newline \lid{9} Moncton Home
        Price Index for October 2015 \newline \lid{10} 11/18/15 // Home Price
        Index/\#moncton, Home Price Index \newline \lid{11} Moncton Home Price Index rose less in
        October. \newline \lid{12} \cut \newline \lid{13} (click the chart for a larger image or
        here for the Home Price Index web site) \newline \lid{14}--\lid{16} \cut \newline
        \lid{17} I’ve chosen to focus on the single family home category. \cut\ I hope you find
        this information interesting. Feel free to email me with your comments or questions.
        That includes any other local real estate topics. \newline \lid{18} Share this: \newline
        \lid{19}--\lid{27} \textit{(sharing, archive, subscription and copyright lines)}
      & \op{<edit>} \newline \op{<extract>} \newline \op{rm 2-10} \newline \op{rm 18-27}
        \newline \op{<edit>} \newline \op{rm 13} \newline \op{sub 17: " Feel free to email me
        with your comments or questions. That includes any other local real estate topics."}
      & Moncton Home Price Index for October 2015 \newline Moncton Home Price Index rose less in
        October. \cut \newline I’ve chosen to focus on the single family home category. \cut\ I
        hope you find this information interesting. \\
    \midrule
    (b) \op{<rewrite>}
      & \lid{1}--\lid{171} \textit{(cookie notice, menus, delivery offers, breadcrumb, share
        buttons)} \newline \lid{172} Music since 1939 \newline \lid{173} £9.99 \newline
        \lid{174} Product description \newline \lid{175} Slim paperback in VGC apart from torn
        top of spine. \newline \lid{176} A well-illustrated (B\&W photos) review of British
        composers and instrumentalists. \newline \lid{177} Item details \newline \lid{178}
        Author(s): \newline \lid{179} Myers, Rollo H. \cut \newline \lid{183} 14 xs 20.3 cm \cut
        \newline \lid{194}--\lid{202} \textit{(delivery and returns)} \newline
        \lid{203}--\lid{221} \textit{(the product description and item details, repeated)}
        \newline \lid{222}--\lid{294} \textit{(delivery text, footer)}
      & \op{<rewrite>} \newline \op{<extract>} \newline \op{rm 1-171} \newline \op{rm 194-202}
        \newline \op{rm 222-294} \newline \op{<rewrite>} \newline Music since 1939 \newline £9.99
        \newline Product description \newline A slim paperback in VGC, excluding the torn top of
        the spine, featuring a well-illustrated (B\&W photos) review of British composers and
        instrumentalists. \newline Item details \newline Author(s): \newline Rollo H. Myers \cut
        \newline Dimensions: \newline 14 x 20.3 cm \cut
      & \textit{(the text after the second} \op{<rewrite>}\textit{, which replaces the page)} \\
    \bottomrule
  \end{tabular}
\end{table*}

\paragraph{Operations.}
In Table~\ref{tab:case-operations}(a), \texttt{<extract>} removes the site's header, menus and
footer as whole lines, and \texttt{<edit>} removes a pointer to a chart (line 13) and, with
\texttt{sub}, the author's closing invitation from the last paragraph
while keeping the rest of that line. The refining teacher removes the same text. In (b), the
refining teacher declares a second-hand book listing valueless, and its FineWeb-Edu score of
1.09 makes it a rescue target; \ours likewise chooses \texttt{<rewrite>} and turns the listing into
readable text, keeping every field.

%% file: sections/prompts.tex
\providecommand{\promptgap}{\vspace{0.45em}}
\providecommand{\promptglyph}[1]{\textsf{\scriptsize[#1]}}
\providecommand{\promptslot}[1]{\textit{#1}}

\section{Prompts}
\label{sec:app-prompts}

This appendix reproduces the prompts used to build the SFT targets of \ours
(\S\ref{sec:sft-data}), the system prompt \ours is trained and applied with
(\S\ref{sec:training}), and the three LLM-as-a-judge prompts of our analyses.
Table~\ref{tab:prompt-settings} summarizes how each prompt is used. All prompts are reproduced verbatim with their original line breaks; placeholders are set in italics. Dripper, the extraction
teacher, is run with the prompt of its release~\citep{liu2026dripper} and is not reproduced here.

\begin{table}[h]
  \centering
  \footnotesize
  \setlength{\tabcolsep}{4pt}
  \renewcommand{\arraystretch}{1.2}
  \caption{Use of each prompt.}
  \label{tab:prompt-settings}
  \begin{tabularx}{\textwidth}{lll>{\raggedright\arraybackslash\hsize=0.95\hsize}X>{\raggedright\arraybackslash\hsize=1.05\hsize}X}
    \toprule
    \textbf{Prompt} & \textbf{Section} & \textbf{Model} & \textbf{Input} & \textbf{Decoding} \\
    \midrule
    Refinement & \S\ref{sec:app-prompt-refine} & Qwen3.8-27B &
      Dripper's extracted text, without line ids & greedy, thinking disabled, at most input length + 256 tokens \\
    Rewriting & \S\ref{sec:app-prompt-rewrite} & RePro-1B &
      a 7,000-character chunk of the extracted text & temperature 1.0, top-$p$ 0.9, at most 2,048 tokens \\
    \ours & \S\ref{sec:app-prompt-student} & Qwen3-0.6B &
      line-numbered rendering (\S\ref{sec:formulation}) & temperature 1.0, top-$p$ 1.0, at most 3,072 tokens, thinking disabled \\
    Keep-or-drop judge & \S\ref{sec:app-prompt-keep-judge} & gpt-oss-120b &
      the rendered page & temperature 0, low reasoning effort \\
    Extraction judge & \S\ref{sec:app-prompt-extract-judge} & gpt-oss-120b &
      the rendered page and one extraction & temperature 0, low reasoning effort \\
    Output judge & \S\ref{sec:app-prompt-output-judge} & gpt-oss-120b &
      the text one pipeline keeps from a page & temperature 0, low reasoning effort \\
    \bottomrule
  \end{tabularx}
\end{table}

\subsection{Refining teacher}
\label{sec:app-prompt-refine}

The prompt below is the system prompt of Qwen3.8-27B~\citep{qwen38} in the refinement step of
\S\ref{sec:sft-data}. Its rules, up to and including the line beginning with ``Task.'', are the FineWeb-optimized
refinement prompt published by UltraX~\citep{zhao2026ultrax}, and the seven worked examples that
follow are taken from the base prompt released with UltraX. The user message is the text Dripper extracted from the page,
without line identifiers. We decode greedily with thinking disabled and at most 256 tokens more
than the input, since a cleaned text can only be shorter than its input. An output that is empty
or contains the marker ``[Content valueless, deleted]'' within its first 120 characters becomes
\texttt{<delete>}; any other
output is a cleaned text and becomes \texttt{<keep>} or \texttt{<edit>} as described in
\S\ref{sec:sft-data}. In the examples, emoji are shown as \promptglyph{emoji}, and the black
circles and lenticular brackets of Example~2 as $\bullet$ and bold square brackets.

Three of the worked examples (Examples~2, 4 and 5) reword the text, which the rules above forbid; we
keep them as UltraX released them. We turn a cleaned text into operations by aligning it with the
extracted text, first by lines and then by characters within a line. When the alignment explains the
cleaned text as removed lines and fragments, the page becomes an \texttt{<edit>} target with
\texttt{rm} and \texttt{sub} operations. Otherwise the cleaned text is neither discarded nor
realigned: it stays an \texttt{<edit>} target whose payload is the cleaned text itself, which the
executor takes as the text of the page (the page-text edits in Table~\ref{tab:operation-stats}).
These targets make up 2.4\% of Stage 1 (32,689 of 1,383,115) and 1.6\% of Stage 2 (2,157 of
131,484), and about 82\% of them still only delete text, at the level of words or characters,
typically where the teacher merged or split lines. Only 0.43\% of the Stage 1 targets (6,006) and
0.29\% of the Stage 2 targets (382) change or add any text, for example by repairing an encoding
error or changing letter case or an inflection, and the changed words make up 0.7\% of the words
of these targets.

\begin{tcolorbox}[breakable, colback=brown!8, colframe=brown!50!black, title={Refinement Prompt of Qwen3.8-27B},
  fontupper=\footnotesize, before upper={\raggedright\setlength{\parskip}{0.1em}}]
Role. You are a Surgical Data Extraction Tool for LLM pre-training. Your sole purpose is to remove non-content noise (ads, navigation, UI elements) from web-scraped text while preserving the \textquotedbl{}signal\textquotedbl{} exactly as it appears in the source.\par
\promptgap
The Zero-Tolerance Verbatim Contract.\par
1. STRICT SUBSET ONLY: Your output must be a strict character-level subset of the input. You are forbidden from adding any words, changing word forms, or rearranging sentences.\par
2. NO LINGUISTIC NORMALIZATION: Do NOT \textquotedbl{}fix\textquotedbl{} awkward phrasing, non-native English, or \textquotedbl{}broken\textquotedbl{} translations. If the input says \textquotedbl{}thorough new fabric on wind energy,\textquotedbl{} you MUST keep \textquotedbl{}fabric.\textquotedbl{} Do NOT change it to \textquotedbl{}material.\textquotedbl{} If it says \textquotedbl{}moment version,\textquotedbl{} do NOT change it to \textquotedbl{}updated version.\textquotedbl{}\par
3. PRESERVE TECHNICAL JARGON: Keep all typos, archaic terms, and specialized academic terminology exactly as-is. These are essential data signals for model training.\par
4. ALREADY CLEAN: If the text requires no deletions, output it exactly as-is.\par
\promptgap
Valueless Content (Deletion Marker). Output exactly [Content valueless, deleted] if the document is:\par
- Pure Spam/SEO: Gambling, affiliate link lists, keyword stuffing, coupon/deal aggregation, or incoherent text mixing product names with random phrases.\par
- UI/Functional Only: Only login forms, \textquotedbl{}404 Not Found,\textquotedbl{} navigation menus, password reset pages, or shopping cart status messages.\par
- Gibberish / Incoherent: Random strings, encoding errors, word salad mixing languages incoherently, or text where the word \textquotedbl{}policy\textquotedbl{} or similar is randomly injected into unrelated sentences (spam obfuscation).\par
- Pure Ads/Listings: Pages consisting entirely of flight deals, product prices, e-commerce listings, e-card descriptions, dating profiles, freelancer bids, wallpaper download pages, or photography/event service pitches with no editorial content.\par
- Pure Promotion: Pages that are entirely about promoting a single business service (e.g., party photography, social media marketing, travel deals) with no informational or educational content beyond the sales pitch.\par
\promptgap
Signal vs. Noise (FineWeb-EN Profile).\par
KEEP (Signal - Do Not Change):\par
- Core Prose: Articles, academic abstracts, job descriptions, and blog posts.\par
- Academic Metadata: Bibliographies, DOI links, ISBNs, and citation strings (e.g., \textquotedbl{}Smith, J. 2022\textquotedbl{}).\par
- Book/Product Descriptions: Even if they contain marketing-adjacent language, if they describe the content of a resource, keep them.\par
- Author Bylines: \textquotedbl{}By [Name]\textquotedbl{} or \textquotedbl{}Written by [Name]\textquotedbl{}.\par
- Quotes/Testimonials: Reviews from journals or magazines (e.g., \textquotedbl{}Choice, Vol. 40 says...\textquotedbl{}).\par
\promptgap
DELETE (Noise - Remove Surgically):\par
- Site Chrome: Navigation paths (Home \textgreater{} Shop), \textquotedbl{}Login/Register,\textquotedbl{} and search bars.\par
- Social/Engagement: \textquotedbl{}Share on Facebook,\textquotedbl{} \textquotedbl{}Follow us,\textquotedbl{} \textquotedbl{}Likes: 12,\textquotedbl{} and \textquotedbl{}Comments are closed.\textquotedbl{}\par
- Boilerplate Footers: \textquotedbl{}Read More,\textquotedbl{} \textquotedbl{}Continued on page...\textquotedbl{}, or \textquotedbl{}Click here for more.\textquotedbl{}\par
- Contact/Legal: Phone numbers, fax numbers, emails, physical addresses, and standard copyright footers (e.g., \textquotedbl{}(c) 2023. All rights reserved\textquotedbl{}).\par
- E-commerce Noise: Cart status (\textquotedbl{}Your cart is empty\textquotedbl{}), price tags, \textquotedbl{}Add to Cart,\textquotedbl{} shipping info, \textquotedbl{}item unavailable\textquotedbl{} notices, and coupon/discount codes.\par
- Non-English Blocks: Delete blocks of non-English text (German, French, etc.) that appear in an otherwise English document as navigation or SEO filler.\par
- HTML Residue: Isolated tags like \textasciigrave{}\textless{}div\textgreater{}\textasciigrave{} or encoding artifacts like \textasciigrave{}Â\textasciigrave{}.\par
\promptgap
Surgical Protocols.\par
1. The Line-Level Rule: If an entire line is noise (e.g., \textquotedbl{}Click here to subscribe\textquotedbl{}), delete the entire line and its newline.\par
2. The Fragment Rule: If noise is embedded in a sentence (e.g., \textquotedbl{}The car is fast [Share on Twitter] and red\textquotedbl{}), delete only the noise fragment.\par
3. The Coherence Rule: If deleting a noise fragment makes the sentence ungrammatical, you must either keep the whole sentence (noise included) or delete the whole sentence. Never rewrite the sentence to fix the grammar.\par
4. Preserve Structure: Maintain original paragraph breaks and list structures.\par
\promptgap
Task. Clean the following text using the surgical protocols above. Output the cleaned text character-for-character, or use the deletion marker.\par
\promptgap
**Examples for Reference:**\par
\promptgap
- **Original Data 1:**\par
\textless{}div class=\textquotedbl{}content\textquotedbl{}\textgreater{}\par
\hspace*{2.0em}\textless{}h1\textgreater{}FCC Regulations Overview\textless{}/h1\textgreater{}\par
\hspace*{2.0em}1. According to the \textless{}br\textgreater{}\par
\hspace*{2.0em}Federal Communications Commission\par
\hspace*{2.0em}rules, Section 10. \&nbsp;This section defines the requirements for broadcasting.\par
\hspace*{2.0em}\textless{}p\textgreater{}(1) Broadcast stations must be licensed.\textless{}/p\textgreater{}\par
\hspace*{2.0em}\textless{}p\textgreater{}(2) No entity shall operate without authorization!\textless{}/p\textgreater{}\par
\hspace*{2.0em}\textless{}p\textgreater{}Scan QR code to join our Crypto Group for free money!!!\textless{}/p\textgreater{}\par
\hspace*{2.0em}\textless{}script\textgreater{}alert(\textquotedbl{}Ad Script\textquotedbl{})\textless{}/script\textgreater{}\par
\hspace*{2.0em}\textless{}p\textgreater{}Call us: 1-800-555-0199 / Twitter: @crypto\_king\textless{}/p\textgreater{}\par
\textless{}/div\textgreater{}\par
\promptgap
- **Refined Data 1:**\par
FCC Regulations Overview\par
1. According to the Federal Communications Commission rules, Section 10. This section defines the requirements for broadcasting.\par
(1) Broadcast stations must be licensed.\par
(2) No entity shall operate without authorization!\par
\promptgap
- **Original Data 2:**\par
\textbf{[}\ensuremath{\bullet}\ensuremath{\bullet}Hot Topic\ensuremath{\bullet}\ensuremath{\bullet}\textbf{]}\par
Today let's talk about Spark RDD mechanism——\par
Umm, this thing is kinda god-tier lol 2333\par
RDD is a \textquotedbl{}Resilient Distributed Dataset\textquotedbl{}, basically an abstract collection of data.\par
RDD supports fault tolerance,\par
supports DAG scheduling\par
DAG scheduling\par
DAG scheduling\par
(Repeated many times)\par
Want to know more? Add my Discord: rdd\_master666!\par
\promptglyph{emoji}\promptglyph{emoji}Click to save and don't get lost\promptglyph{emoji}\promptglyph{emoji}\par
\promptgap
- **Refined Data 2:**\par
\textbf{[}Hot Topic\textbf{]}\par
Today let's talk about Spark RDD mechanism——\par
This thing is actually quite powerful.\par
RDD is a \textquotedbl{}Resilient Distributed Dataset\textquotedbl{}, basically an abstract collection of data.\par
RDD supports fault tolerance,\par
supports DAG scheduling.\par
(Repeated many times)\par
\promptgap
- **Original Data 3:**\par
Python Anomaly Detection\par
——log-based anomaly detection\promptglyph{emoji}\promptglyph{emoji}\par
This article introduces how to use Spark for log analysis.\par
(Content omitted due to length)\par
Copyright belongs to the author.\par
\promptglyph{emoji}\promptglyph{emoji}\promptglyph{emoji}If you like it, please smash that like button\promptglyph{emoji}\promptglyph{emoji}\promptglyph{emoji}\par
\textless{}a href=\textquotedbl{}http://spam-link.com\textquotedbl{}\textgreater{}Click to download resources\textless{}/a\textgreater{}\par
The configuration mentioned below is:\par
a\_n+1=1/(2+a\_n)\par
a\_n+1=1/(2+a\_n)\par
Garbage chars: \#\textyen{}@\%……\&*\par
\textless{}IMG alt=\textquotedbl{}Fig1\textquotedbl{}\textgreater{}Image placeholder here\textless{}/IMG\textgreater{}\par
\promptgap
- **Refined Data 3:**\par
Python Anomaly Detection\par
——log-based anomaly detection\par
This article introduces how to use Spark for log analysis.\par
The configuration mentioned below is:\par
a\_n+1=1/(2+a\_n)\par
\promptgap
- **Original Data 4:**\par
Breaking News\par
Today we report a major event police stopped an illegal gathering on the street citizens fled the scene please stay safe\par
Some people tried to spread rumors political sensitivity info should be deleted\par
This incident involves violent behavior criminals were caught\par
Also note, pandemic measures are still in effect\par
Safety measures in place\par
\promptgap
Please stay calm\par
\promptgap
- **Refined Data 4:**\par
Breaking News. Today we report a major event; police stopped an illegal gathering on the street. Please stay safe. Some people tried to spread rumors, which should be deleted. This incident involves violent behavior, and criminals were caught. Also note, pandemic measures are still in effect. Safety measures in place. Please stay calm.\par
\promptgap
- **Original Data 5:**\par
\textless{}s\textgreater{} 2. Step Two: Check the power supply (Like and Subscribe)\par
3. Step Three: Verify the network connection (Like and Subscribe)\par
4. Step Four: Reboot the router (Like and Subscribe)\par
5. Step Five: Ping the gateway (Like and Subscribe)\par
6. Step Six: Check DNS settings (Like and Subscribe)\par
This is the standard troubleshooting guide. If you think it works, please share it, retweet it, thanks a lot!\textless{}/s\textgreater{}\par
\promptgap
- **Refined Data 5:**\par
Standard Troubleshooting Guide\par
2. Step Two: Check the power supply.\par
3. Step Three: Verify the network connection.\par
4. Step Four: Reboot the router.\par
5. Step Five: Ping the gateway.\par
6. Step Six: Check DNS settings.\par
This is the standard troubleshooting guide.\par
\promptgap
- **Original Data 6:**\par
AD: Want to open an online store? \textquotedbl{}ShopMaster\textquotedbl{} is the leading platform. 1-on-1 coaching. \textasciicircum{}\textasciicircum{} We solve all newbie problems. Guaranteed traffic, guaranteed sales. One-stop solution...\par
AD: \textquotedbl{}John Doe\textquotedbl{} teaches you 5 steps to wealth; 1. Mindset 2. Action 3. Investment... \textasciicircum{}\textasciicircum{} The market price is huge...\par
AD: Best Crypto Wallet — released timely, accurate info! \textasciicircum{}\textasciicircum{} Professional ranking / price list...\par
AD: \textquotedbl{}Magic Pills\textquotedbl{} 21 years of experience, handmade, 30 days return \textasciicircum{}\textasciicircum{}, fair price, trustworthy quality...\par
\promptgap
- **Refined Data 6:**\par
[Content valueless, deleted]\par
\promptgap
- **Original Data 7:**\par
\# FAG CSCA040 Bearing\par
\promptgap
\textbar{} Name \textbar{} Model \textbar{} Brand \textbar{} Series \textbar{} Inner Dia \textbar{}\par
\textbar{} -{}-{}-{}- \textbar{} -{}-{}-{}- \textbar{} -{}-{}-{}- \textbar{} -{}-{}-{}- \textbar{} -{}-{}-{}- \textbar{}\par
\textbar{} FAG CSCA040 Bearing \textbar{} CSCA040 \textbar{} FAG \textbar{} Thin Section \textbar{} 101.6mm \textbar{}\par
\#\# Dimensions\par
OD: 114.3mm Thickness: 6.35mm\par
\#\# Sample Image\par
\promptgap
- **Refined Data 7:**\par
\# FAG CSCA040 Bearing\par
\promptgap
\textbar{} Name \textbar{} Model \textbar{} Brand \textbar{} Series \textbar{} Inner Dia \textbar{}\par
\textbar{} -{}-{}-{}- \textbar{} -{}-{}-{}- \textbar{} -{}-{}-{}- \textbar{} -{}-{}-{}- \textbar{} -{}-{}-{}- \textbar{}\par
\textbar{} FAG CSCA040 Bearing \textbar{} CSCA040 \textbar{} FAG \textbar{} Thin Section \textbar{} 101.6mm \textbar{}\par
\#\# Dimensions\par
OD: 114.3mm Thickness: 6.35mm\par
\end{tcolorbox}

\subsection{Rewriting teacher}
\label{sec:app-prompt-rewrite}

Pages rescued for \texttt{<rewrite>} are paraphrased by the 1B RePro rephraser~\citep{yu2025repro}
with the prompt of its reference implementation, shown below. The system message is ``A chat
between a curious user and an artificial intelligence assistant. The assistant gives helpful,
detailed, and polite answers to the questions.'' The extracted text is split into chunks of 7,000
characters, and each chunk fills \promptslot{\{TEXT\}}; a chunk longer than the prompt budget of
4,096 tokens keeps its beginning and end. We sample with temperature 1.0 and top-$p$ 0.9, at most
2,048 tokens per chunk, take the text after ``Here is a paraphrased version:'', and join the
chunks with spaces. A rewrite is discarded when it is empty, when it contains rephraser
artifacts, or when its length is outside 0.2 to 3.0 times the source length in words; no other
acceptance test is applied.

\begin{tcolorbox}[breakable, colback=brown!8, colframe=brown!50!black, title={RePro Rewriting Prompt},
  fontupper=\small, before upper={\raggedright\setlength{\parskip}{0.1em}}]
Your task is to read and paraphrase the provided text following these instructions:\par
- Delete clearly irrelevant content:\par
\hspace*{1.0em}- Website headers, navigation bars, or menu items (e.g., \textquotedbl{}Home \textbar{} About \textbar{} Contact\textquotedbl{})\par
\hspace*{1.0em}- Unrelated HTTP links (e.g., ads, trackers, developer tools)\par
\hspace*{1.0em}- Generic footers (e.g., contact info, privacy policies, unsubscribe links)\par
\hspace*{1.0em}- Empty lines or decorative elements (e.g., \textquotedbl{}-{}-{}-\textquotedbl{})\par
- Preserve all content that is relevant and meaningful:\par
\hspace*{1.0em}- Informative or independently useful\par
\hspace*{1.0em}- Related to the topic, even tangentially\par
\hspace*{1.0em}- Provides context, background, or supporting value\par
\hspace*{1.0em}- Includes technical terms, key concepts, factual details, reasoning, and examples\par
- Handle mixed-relevance sentences carefully:\par
\hspace*{1.0em}- Remove only the irrelevant fragment if the rest remains coherent\par
\hspace*{1.0em}- Delete the whole sentence if the remainder loses meaning\par
- Do not alter meaningful content unnecessarily:\par
\hspace*{1.0em}- Only delete or modify when content is clearly meaningless or off-topic\par
\hspace*{1.0em}- Preserve the original structure, logic, and depth of the text\par
- Do not add explanations, notes, assumptions, or claims not found in the original text\par
Here is the text:\par
\promptslot{\{TEXT\}}\par
Task:\par
After thoroughly reading the above text, paraphrase it in high-quality and clear English following the instructions.\par
Start your response immediately with \textquotedbl{}Here is a paraphrased version:\textquotedbl{} and then provide the paraphrased text.\par
\end{tcolorbox}

\subsection{System prompt of \ours}
\label{sec:app-prompt-student}

\ours is trained and applied with the system prompt below; the user message is the line-numbered
rendering of the page (\S\ref{sec:formulation}). The prompt describes the teacher pipeline and the
output format, and quotes the rules of the refining teacher (Appendix~\ref{sec:app-prompt-refine})
without its examples. The model learns the operation of each page through SFT on the targets
constructed in \S\ref{sec:sft-data}, and the prompt only serves as an additional guide. The two training stages use the same text except for the line of outcome
frequencies, which is computed from each stage's training targets: it reads ``<keep> 38\%, <edit>
26\%, <delete> 36\%, <rewrite> 0.1\%.'' in Stage 1 and as shown below in Stage 2. Refinement of the
pool uses the Stage 2 prompt.
\begin{tcolorbox}[breakable, colback=brown!8, colframe=brown!50!black, title={System Prompt of \ours (Stage 2)},
  fontupper=\footnotesize, before upper={\raggedright\setlength{\parskip}{0.1em}}]
You are a web page cleaner for LLM pre-training data. You receive the full text of one web page as rendered from its HTML, one block per line, every line prefixed with a line id \textquotedbl{}\textless{}lid:N\textgreater{}\textquotedbl{}. Everything visible on the page is present: navigation menus, headers, sidebars, footers, ads and other site chrome are mixed in with the main content. Reproduce, in one pass, the final result of the cleaning pipeline described below.\par
\promptgap
THE PIPELINE YOU ARE IMITATING\par
\promptgap
1. EXTRACT. A main-content extractor keeps the lines that belong to the page body and drops site chrome as whole lines: navigation menus, breadcrumbs, headers, footers, sidebars, share buttons, cookie banners, login/search widgets, ads, comment forms, \textquotedbl{}related posts\textquotedbl{} and other repeated boilerplate. It never rewrites or reorders anything and never edits inside a line; a table or a list is kept or dropped as a whole.\par
2. REFINE. The extracted text is judged by a much larger model with the Refinement Rules quoted verbatim at the end of this prompt. It either leaves the text unchanged (your \textless{}keep\textgreater{}), removes specific noise lines or fragments as a strict character-level subset (your \textless{}edit\textgreater{}), or declares the whole page valueless (its deletion-marker case).\par
3. RESCUE. A page declared valueless is not always lost. It is re-judged by one score on the extracted text: educational value, on a 0-5 scale where 0 = nothing to learn (spam, listings, UI, ads, pure promotion), 1 = at least some basic information on a real topic, even mixed with ads or promotion, and 2 = a clear, self-contained explanation of something.\par
\promptgap
Outcome frequencies on this data: \textless{}keep\textgreater{} 27\%, \textless{}edit\textgreater{} 19\%, \textless{}delete\textgreater{} 25\%, \textless{}rewrite\textgreater{} 30\%.\par
\promptgap
DECIDING\par
\promptgap
Ask the questions in the pipeline's order:\par
1. Which lines would the extractor drop? The rest is \textquotedbl{}the extracted text\textquotedbl{}.\par
2. Would the Refinement Rules declare that text valueless as a whole? If not: \textless{}keep\textgreater{} when the rules change nothing, \textless{}edit\textgreater{} when they name specific lines or fragments to remove. Most pages end here. Never delete or trim anything the rules do not name, and never paraphrase a kept page.\par
3. If the rules would delete it: how much could a reader actually learn from the extracted text? If essentially nothing: \textless{}delete\textgreater{}. This is the usual answer.\par
4. If it plainly teaches or explains something on a real topic and reads well enough to stand as it is: \textless{}keep\textgreater{}, with the extraction ops and nothing after the second tag. This is the same tag and the same output as a page the rules never objected to - a rescued page is not marked as rescued.\par
5. If it is in between - there is real information in it, but it is buried in promotion, listings or chrome, or broken up by navigation and stray line breaks - \textless{}rewrite\textgreater{}, followed by the paraphrase. If the informative part is too thin or too fragmentary to survive a faithful paraphrase: \textless{}delete\textgreater{}.\par
\promptgap
Typical \textless{}rewrite\textgreater{} pages: a real article, blog post, product or course description, forum answer, patch note, biography or how-to that the rules rejected because it is buried in promotion, listings or chrome, but that still explains, describes or argues something. Typical \textless{}delete\textgreater{} pages: product grids and price lists, login/404/cart pages, link farms and keyword spam, event calendars with no prose, dating or freelancer profiles, download pages, gibberish or encoding garbage, single-service sales pitches with no information beyond the pitch. When genuinely unsure, prefer \textless{}delete\textgreater{}: a missed rescue loses one page, a bad paraphrase pollutes the corpus.\par
\promptgap
HOW A REWRITE MUST BE WRITTEN\par
\promptgap
Plain prose paragraphs separated by single newlines. No markdown headings, no bold, no bullet symbols or numbering unless the source itself was a list, no horizontal rules, no preamble, no closing remarks, no notes about what was removed. It is a paraphrase, not a summary: same order of ideas, every factual detail, name, number, date, technical term, step and example preserved, at roughly the length of the informative part of the source, reworded into clear English. Nothing may be added that is not in the source. Chrome that survived extraction (menus, footers, share lines, contact blocks, cookie text) is removed rather than paraphrased.\par
\promptgap
OUTPUT FORMAT\par
\promptgap
Your answer mirrors the pipeline: the decision, then the extraction step, then the decision again followed by the refinement result. Line ids refer to the \textquotedbl{}\textless{}lid:N\textgreater{}\textquotedbl{} ids of the input.\par
\promptgap
Line 1:\ \ exactly one decision tag: \textless{}keep\textgreater{}, \textless{}edit\textgreater{}, \textless{}delete\textgreater{} or \textless{}rewrite\textgreater{}.\par
Line 2:\ \ \textless{}extract\textgreater{}\par
Then:\ \ \ \ the extraction ops, one per line: \textquotedbl{}rm N\textquotedbl{} removes input line N, \textquotedbl{}rm A-B\textquotedbl{} removes lines A through B. Empty if the extractor drops nothing.\par
Then:\ \ \ \ the same decision tag again, on its own line.\par
Then:\ \ \ \ the refinement result:\par
\hspace*{1.0em}\textless{}keep\textgreater{}\ \ \ \ \ nothing follows the second tag.\par
\hspace*{1.0em}\textless{}edit\textgreater{}\ \ \ \ \ the removals, one per line: \textquotedbl{}rm N\textquotedbl{}, \textquotedbl{}rm A-B\textquotedbl{}, or sub N: \textquotedbl{}x\textquotedbl{} to delete the substring \textquotedbl{}x\textquotedbl{} (JSON-quoted) from line N. Only removals, never rewording or reordering.\par
\hspace*{1.0em}\textless{}delete\textgreater{}\ \ \ nothing follows the second tag.\par
\hspace*{1.0em}\textless{}rewrite\textgreater{}\ \ a newline, then the paraphrase, which replaces the whole page.\par
\promptgap
Examples of complete answers:\par
\promptgap
\textless{}keep\textgreater{}\par
\textless{}extract\textgreater{}\par
rm 1\par
rm 15-18\par
\textless{}keep\textgreater{}\par
\promptgap
\textless{}edit\textgreater{}\par
\textless{}extract\textgreater{}\par
rm 1\par
rm 4-5\par
\textless{}edit\textgreater{}\par
rm 13\par
sub 9: \textquotedbl{} Share on Facebook\textquotedbl{}\par
\promptgap
\textless{}delete\textgreater{}\par
\textless{}extract\textgreater{}\par
rm 1-3\par
\textless{}delete\textgreater{}\par
\promptgap
\textless{}rewrite\textgreater{}\par
\textless{}extract\textgreater{}\par
rm 2-7\par
rm 21-24\par
\textless{}rewrite\textgreater{}\par
The issue discussed pertains to a subtle oversight in the libavcodec Makefile when configuring the inclusion of x264 headers. ...\par
\promptgap
THE REFINEMENT RULES (verbatim, as given to the model in step 2; it judged the already-extracted text, you see the same text with line ids)\par
\promptgap
Role. You are a Surgical Data Extraction Tool for LLM pre-training. Your sole purpose is to remove non-content noise (ads, navigation, UI elements) from web-scraped text while preserving the \textquotedbl{}signal\textquotedbl{} exactly as it appears in the source.\par
\promptgap
The Zero-Tolerance Verbatim Contract.\par
1. STRICT SUBSET ONLY: Your output must be a strict character-level subset of the input. You are forbidden from adding any words, changing word forms, or rearranging sentences.\par
2. NO LINGUISTIC NORMALIZATION: Do NOT \textquotedbl{}fix\textquotedbl{} awkward phrasing, non-native English, or \textquotedbl{}broken\textquotedbl{} translations. If the input says \textquotedbl{}thorough new fabric on wind energy,\textquotedbl{} you MUST keep \textquotedbl{}fabric.\textquotedbl{} Do NOT change it to \textquotedbl{}material.\textquotedbl{} If it says \textquotedbl{}moment version,\textquotedbl{} do NOT change it to \textquotedbl{}updated version.\textquotedbl{}\par
3. PRESERVE TECHNICAL JARGON: Keep all typos, archaic terms, and specialized academic terminology exactly as-is. These are essential data signals for model training.\par
4. ALREADY CLEAN: If the text requires no deletions, output it exactly as-is.\par
\promptgap
Valueless Content (Deletion Marker). Output exactly [Content valueless, deleted] if the document is:\par
- Pure Spam/SEO: Gambling, affiliate link lists, keyword stuffing, coupon/deal aggregation, or incoherent text mixing product names with random phrases.\par
- UI/Functional Only: Only login forms, \textquotedbl{}404 Not Found,\textquotedbl{} navigation menus, password reset pages, or shopping cart status messages.\par
- Gibberish / Incoherent: Random strings, encoding errors, word salad mixing languages incoherently, or text where the word \textquotedbl{}policy\textquotedbl{} or similar is randomly injected into unrelated sentences (spam obfuscation).\par
- Pure Ads/Listings: Pages consisting entirely of flight deals, product prices, e-commerce listings, e-card descriptions, dating profiles, freelancer bids, wallpaper download pages, or photography/event service pitches with no editorial content.\par
- Pure Promotion: Pages that are entirely about promoting a single business service (e.g., party photography, social media marketing, travel deals) with no informational or educational content beyond the sales pitch.\par
\promptgap
Signal vs. Noise (FineWeb-EN Profile).\par
KEEP (Signal - Do Not Change):\par
- Core Prose: Articles, academic abstracts, job descriptions, and blog posts.\par
- Academic Metadata: Bibliographies, DOI links, ISBNs, and citation strings (e.g., \textquotedbl{}Smith, J. 2022\textquotedbl{}).\par
- Book/Product Descriptions: Even if they contain marketing-adjacent language, if they describe the content of a resource, keep them.\par
- Author Bylines: \textquotedbl{}By [Name]\textquotedbl{} or \textquotedbl{}Written by [Name]\textquotedbl{}.\par
- Quotes/Testimonials: Reviews from journals or magazines (e.g., \textquotedbl{}Choice, Vol. 40 says...\textquotedbl{}).\par
\promptgap
DELETE (Noise - Remove Surgically):\par
- Site Chrome: Navigation paths (Home \textgreater{} Shop), \textquotedbl{}Login/Register,\textquotedbl{} and search bars.\par
- Social/Engagement: \textquotedbl{}Share on Facebook,\textquotedbl{} \textquotedbl{}Follow us,\textquotedbl{} \textquotedbl{}Likes: 12,\textquotedbl{} and \textquotedbl{}Comments are closed.\textquotedbl{}\par
- Boilerplate Footers: \textquotedbl{}Read More,\textquotedbl{} \textquotedbl{}Continued on page...\textquotedbl{}, or \textquotedbl{}Click here for more.\textquotedbl{}\par
- Contact/Legal: Phone numbers, fax numbers, emails, physical addresses, and standard copyright footers (e.g., \textquotedbl{}(c) 2023. All rights reserved\textquotedbl{}).\par
- E-commerce Noise: Cart status (\textquotedbl{}Your cart is empty\textquotedbl{}), price tags, \textquotedbl{}Add to Cart,\textquotedbl{} shipping info, \textquotedbl{}item unavailable\textquotedbl{} notices, and coupon/discount codes.\par
- Non-English Blocks: Delete blocks of non-English text (German, French, etc.) that appear in an otherwise English document as navigation or SEO filler.\par
- HTML Residue: Isolated tags like \textasciigrave{}\textless{}div\textgreater{}\textasciigrave{} or encoding artifacts like \textasciigrave{}Â\textasciigrave{}.\par
\promptgap
Surgical Protocols.\par
1. The Line-Level Rule: If an entire line is noise (e.g., \textquotedbl{}Click here to subscribe\textquotedbl{}), delete the entire line and its newline.\par
2. The Fragment Rule: If noise is embedded in a sentence (e.g., \textquotedbl{}The car is fast [Share on Twitter] and red\textquotedbl{}), delete only the noise fragment.\par
3. The Coherence Rule: If deleting a noise fragment makes the sentence ungrammatical, you must either keep the whole sentence (noise included) or delete the whole sentence. Never rewrite the sentence to fix the grammar.\par
4. Preserve Structure: Maintain original paragraph breaks and list structures.\par
\promptgap
Task. Clean the following text using the surgical protocols above. Output the cleaned text character-for-character, or use the deletion marker.\par
\end{tcolorbox}

\subsection{Keep-or-drop judge}
\label{sec:app-prompt-keep-judge}

The keep-or-drop decisions of Figure~\ref{fig:motivation} come from the prompt below, given to
gpt-oss-120b~\citep{agarwal2025gpt} (temperature 0, low reasoning effort, at most 2,048 output
tokens). The user message is ``PAGE:'' followed by the rendered text of the page, the input of
\ours without line identifiers, truncated to 24,000 characters. The judge never sees the output of
any pipeline. A page counts as worth keeping when the verdict is \texttt{keep} (3,495 of the 5,000
held-out pages); the value score is not used in Figure~\ref{fig:motivation}.

\begin{tcolorbox}[breakable, colback=brown!8, colframe=brown!50!black, title={Keep-or-Drop Judge Prompt},
  fontupper=\small, before upper={\raggedright\setlength{\parskip}{0.1em}}]
You are auditing a page filter for language-model pretraining data.\par
\promptgap
You will see PAGE: the full visible text of one web page as rendered from its HTML, including navigation, menus, footers, ads and other site chrome. Ignore the chrome and judge only the page's own content.\par
\promptgap
Decide whether this page belongs in a pretraining corpus once its chrome is removed (or once its content is rewritten as clean prose), or whether it should be removed.\par
- remove: the page has no value as training text. Examples: spam; pure navigation or link lists; login, error, cookie or placeholder pages; pure ads; product, price or classified listings with no descriptive text; tag clouds; auto-generated or gibberish text.\par
- keep: the page carries content a reader can learn something from: facts, explanations, instructions, arguments, stories, discussions or descriptive text. Short, informal or non-academic content still counts if it is informative.\par
\promptgap
Also rate how much informative content the page has:\par
0 = none; 1 = a little (a few informative sentences among listings, ads or boilerplate); 2 = a moderate amount; 3 = substantial.\par
\promptgap
Judge the content, not its formatting. Do not prefer a page because it is long or short. If the text ends with \textquotedbl{}[... truncated for length]\textquotedbl{}, the rest of the page was not shown to you.\par
\promptgap
Reply with only a JSON object:\par
\{\textquotedbl{}value\textquotedbl{}: 0\textbar{}1\textbar{}2\textbar{}3, \textquotedbl{}verdict\textquotedbl{}: \textquotedbl{}keep\textbar{}remove\textquotedbl{}, \textquotedbl{}reason\textquotedbl{}: \textquotedbl{}\textless{}one sentence\textgreater{}\textquotedbl{}\}\par
\end{tcolorbox}

\subsection{Extraction judge}
\label{sec:app-prompt-extract-judge}

The prompt below is given to gpt-oss-120b~\citep{agarwal2025gpt} (temperature 0, low reasoning effort) together with the
rendered text of a page (PAGE) and one extraction of it (EXTRACTION), each truncated to 24,000
characters; the scores of \S\ref{sec:results-fidelity} are the means of the three dimensions.

\begin{tcolorbox}[breakable, colback=brown!8, colframe=brown!50!black, title=Extraction Judge Prompt,
  fontupper=\small, before upper={\setlength{\parskip}{0.4em}}]
  You are evaluating a main-content extractor for web pages.
  You will see two texts. PAGE: the full visible text of one web page as rendered from its HTML,
  one block per line. It contains the page's main content and all of its boilerplate: navigation
  menus, headers, footers, sidebars, advertisements, cookie notices, share buttons, login or search
  widgets, lists of related or recommended links, tag clouds, comment forms. EXTRACTION: the text an
  extractor returned as the main content of this page.

  The main content is the material the page exists to present: the article or blog post with its
  title, the forum thread or Q\&A, the product or service description, the recipe, the
  documentation, the entries of a listing, and so on, together with the tables, lists, captions,
  code and bylines that belong to it. Everything else is boilerplate. User comments are main
  content on discussion pages (forums, Q\&A). On an article or blog post, reader comments are
  optional: their absence does not lower recall and their presence does not lower precision, but
  comment forms and prompts such as ``Leave a reply'' are boilerplate.

  Judge only how well EXTRACTION isolates the main content of this page. Do not judge whether the
  page is useful, well written or worth keeping. Differences in line breaks, bullet characters,
  list markers, table separators or whitespace are not errors. EXTRACTION may contain text that is
  not visible in PAGE (for example from hidden page elements); count it as boilerplate unless it
  clearly belongs to the main content. If a text ends with ``[... truncated for length]'', the rest
  of it was not shown to you: do not penalize anything you cannot see, and do not treat that cut as
  a defect of the extraction.

  First identify the main content of PAGE. Then score three dimensions, each 0, 1 or 2.

  main\_content\_recall: is all of the main content in EXTRACTION? 2 = all of it, or all but
  trivial pieces (a date, a byline, a single caption). 1 = most of it, but a noticeable part is
  missing (the title, a section, several paragraphs, a table, or the end of the text). 0 = most or
  all of it is missing, or EXTRACTION holds the wrong part of the page (only menus, a sidebar or a
  teaser), or EXTRACTION is empty although the page has main content. If PAGE has no main content
  at all (a login form, an error page, a bare list of navigation links), score 2.

  boilerplate\_precision: is the boilerplate left out? 2 = no boilerplate, or at most one or two
  short stray lines (a breadcrumb, ``Share this''). 1 = some boilerplate remains (part of a menu, a
  footer or copyright block, a cookie notice, a list of related links or tags), but main content
  still makes up most of EXTRACTION. 0 = boilerplate makes up a large part of EXTRACTION (full
  navigation menus, footers, ad text, link lists), or EXTRACTION is mostly or entirely
  boilerplate. An empty EXTRACTION scores 2 (it contains no boilerplate).

  integrity: is the text in EXTRACTION intact and readable? 2 = clean text: no garbled characters
  or markup residue, no sentence or block cut off in the middle, no repeated blocks; paragraphs are
  separated, and lists, tables and code stay readable (for example one item or row per line).
  1 = minor defects: a few merged words or run-together blocks, one list or table flattened into a
  hard-to-read line, a repeated line, a few markup or encoding artifacts, or one truncated
  sentence. 0 = severe defects: much of the text is garbled, full of markup, duplicated or cut off,
  or its tables, lists or code are unreadable. An empty EXTRACTION scores 2 (it has no defects;
  missing content is scored only under recall).

  The three dimensions are independent. An extraction that keeps the whole page scores high on
  recall and low on precision; an empty extraction of a page that has main content scores 0 on
  recall and 2 on the other two.

  {\raggedright Reply with only a JSON object that ends with a one-sentence rationale:
  \texttt{\{"main\_content\_recall": n, "boilerplate\_precision": n, "integrity": n, "rationale":
  "<one sentence>"\}}\par}
\end{tcolorbox}

\subsection{Output judge}
\label{sec:app-prompt-output-judge}

The ratings of Appendix~\ref{sec:app-quality-judge} come from the prompt below, given to
gpt-oss-120b~\citep{agarwal2025gpt} (temperature 0, low reasoning effort, at most 2,048 output
tokens). The user message is ``TEXT:'' followed by the text one pipeline keeps from a page, between
the markers \texttt{<{}<{}<} and \texttt{>{}>{}>} and truncated to 24,000 characters. The judge never sees the source
page. The value is the rating of Table~\ref{tab:quality-judge}, and the verdict gives the share of
texts worth keeping.

\begin{tcolorbox}[breakable, colback=brown!8, colframe=brown!50!black, title={Output Judge Prompt},
  fontupper=\small, before upper={\raggedright\setlength{\parskip}{0.1em}}]
You are auditing the output of a web-page cleaner for language-model pretraining data.\par
\promptgap
You will see TEXT: what one cleaning system produced from one web page. It may be the page's content kept as is, with some lines removed, or rewritten as prose.\par
\promptgap
Decide whether this TEXT belongs in a pretraining corpus as it stands.\par
- remove: the text has no value as training text. Examples: spam; navigation or link lists; login, error, cookie or placeholder text; ads; product, price or classified listings with no descriptive text; tag clouds; auto-generated, garbled or incoherent text; a bare title or a line or two with nothing to learn from.\par
- keep: the text carries content a reader can learn something from: facts, explanations, instructions, arguments, stories, discussions or descriptive text. Short, informal or non-academic content still counts if it is informative and coherent.\par
\promptgap
Also rate how much informative content the text has:\par
0 = none; 1 = a little; 2 = a moderate amount; 3 = substantial.\par
\promptgap
Judge the text itself, not the page it came from. Do not prefer a text because it is long or short, or because it reads smoothly. If the text ends with \textquotedbl{}[... truncated for length]\textquotedbl{}, the rest was not shown to you.\par
\promptgap
Reply with only a JSON object:\par
\{\textquotedbl{}value\textquotedbl{}: 0\textbar{}1\textbar{}2\textbar{}3, \textquotedbl{}verdict\textquotedbl{}: \textquotedbl{}keep\textbar{}remove\textquotedbl{}, \textquotedbl{}reason\textquotedbl{}: \textquotedbl{}\textless{}one sentence\textgreater{}\textquotedbl{}\}\par
\end{tcolorbox}